\documentclass{article}
\usepackage{iclr2027_conference,times}

\usepackage{amsmath,amsfonts,bm}

\def\eqref#1{equation~\ref{#1}}

\def\1{\bm{1}}

\DeclareMathAlphabet{\mathsfit}{\encodingdefault}{\sfdefault}{m}{sl}
\SetMathAlphabet{\mathsfit}{bold}{\encodingdefault}{\sfdefault}{bx}{n}

\newcommand{\R}{\mathbb{R}}

\usepackage{hyperref}
\usepackage{url}
\usepackage{amsmath,amssymb,amsthm}
\usepackage{bm}
\usepackage{mathtools}
\usepackage{graphicx}
\usepackage{booktabs}
\usepackage{tabularx}
\usepackage{array}
\usepackage{enumitem}
\usepackage{caption}
\usepackage{subcaption}

\newcolumntype{Y}{>{\raggedright\arraybackslash}X}

\newcommand{\bx}{\bm{x}}
\newcommand{\bp}{\bm{p}}
\newcommand{\bb}{\bm{b}}
\newcommand{\cvec}{\bm{c}}
\newcommand{\Real}{\mathrm{Re}}

\newcommand{\lap}{\nabla^2}
\newcommand{\bib}{\nabla^4}
\newcommand{\Gram}{\bm{G}}

\theoremstyle{definition}

\title{Learning Physics from an Imperfect Ancestor}

\author{
S. Mohammad Mousavi$^{1,2}$,
Teeratorn Kadeethum$^{2}$,
Nikolaos Bouklas$^{1}$,
Somdatta Goswami$^{3}$\\[0.5em]
$^{1}$Sibley School of Mechanical and Aerospace Engineering, Cornell University\\
$^{2}$AI Lab, Siemens Energy\\
$^{3}$Department of Civil and Systems Engineering, Johns Hopkins University
}

\iclrfinalcopy

\begin{document}

\maketitle

\begin{abstract}
Neural operators evaluate parametric partial differential equations (PDEs) cheaply but degrade sharply outside their training distribution. Physics-informed neural networks (PINNs) avoid dependence on labeled data, yet their optimization can be basin-fragile: when the governing residual admits multiple solutions, a PINN trained from scratch may converge to a physically incorrect state despite achieving a small residual. We show that these failure modes can be addressed jointly: an imperfect NO provides the structural prior needed to place a PINN in the correct solution basin, while the PDE residual refines the solution beyond the operator's accuracy. We introduce a three-stage framework that freezes the spatial basis of a physics-informed NO, extrapolates its solution branch to an out-of-distribution parameter using a polynomial continuation prior, and distills the resulting field into a fresh PINN. The NO (teacher) need not be accurate at the target; it transfers solution-branch information, while PDE residual minimization in the PINN (student) governs convergence. We evaluate the framework on three nonlinear PDEs: 1D viscous Burgers, 2D steady Allen--Cahn near a pitchfork bifurcation, and 2D steady lid-driven cavity flow. For Allen--Cahn, where the trivial solution satisfies the PDE residual exactly, a standard PINN consistently collapses to the trivial zero branch, whereas distillation from the crude extrapolated operator recovers the non-trivial branch that matches the finite-difference reference. For the lid-driven cavity, extrapolating to a Reynolds number of Re = 3200 accelerates convergence to the correct physical state, achieving competitive accuracy using fewer parameters and optimization steps than recent literature baselines. These results establish a simple principle: an NO need not accurately predict the solution to be useful; it only needs to identify the correct basin from which PINN optimization can recover it.
\end{abstract}

\section{Introduction}
\label{sec:intro}

Parametric partial differential equations (PDEs) underpin many scientific and engineering workflows, where the same governing equations must be solved repeatedly across varying parameters. Neural operators (NOs) amortize this cost by learning parameter-to-solution maps, enabling inexpensive inference after training. However, their accuracy can deteriorate sharply outside the training distribution, while adapting them to new parameters typically requires additional simulations or labeled data. Physics-informed neural networks (PINNs) provide a complementary alternative by enforcing the governing equations directly, but their optimization is sensitive to initialization and can converge to physically incorrect solutions when the PDE residual admits multiple minima. In such cases, a small residual does not guarantee selection of the physically relevant solution branch. These failure modes suggest a complementary division of roles: an NO can provide structural information about the solution manifold, while a PINN can enforce the governing physics at the target parameter. We therefore ask whether an imperfect, out-of-distribution NO can guide a PINN without constraining its final accuracy. We show that it can. The key is to use the NO not as a predictor of the final solution, but as a prior for solution-basin selection.

To this end, we introduce a three-stage framework: a physics-informed NO is trained and its spatial basis frozen; the branch is extrapolated to an out-of-distribution target using a polynomial continuation prior, without any reference solution; and the extrapolated field is distilled into a freshly initialized PINN, whose subsequent residual minimization governs final accuracy. The resulting NO acts as a teacher, but unlike conventional knowledge distillation, it is not required to provide an accurate prediction. Its role is to transfer solution-branch information and place the PINN in a favorable basin of attraction; subsequent minimization of the PDE residual allows the PINN to refine the field and, in principle, surpass the accuracy of the teacher. This separation of roles is central to the proposed framework.

We evaluate the framework on three nonlinear PDEs. First, the 1D viscous Burgers equation establishes the baseline on a problem with a single solution branch. Second, the 2D steady Allen--Cahn equation probes basin selection near a pitchfork bifurcation, a scenario where the trivial solution $u\equiv0$ satisfies the residual exactly. Third, to test substantial out-of-distribution capabilities, we apply the framework to a 2D steady lid-driven cavity, extending an operator trained on a Reynolds number range of $\mathrm{Re}\in[100,800]$ to $\mathrm{Re}=3200$. Finally, an ablation study isolates the underlying mechanism by removing the extrapolation stage, designed to determine whether the critical challenge in these failure modes is strictly residual optimization or physical basin selection.

\section{Related work}
\label{sec:related}

\paragraph{Operator learning.} NOs learn maps between function spaces and are, in principle, discretization-invariant~\citep{kovachki2023neural}. DeepONet~\citep{lu2021learning, haghighat2025stonet} realizes this through a trunk--branch decomposition; the Fourier neural operator~\citep{li2020fourier} and the earlier graph-kernel network~\citep{li2020neural} operate in spectral and graph domains respectively, with Wavelet~\citep{tripura2023wavelet},Laplace~\citep{cao2024laplace} and other variants targeting localized features~\citep{lu2022comprehensive,cao2024deep}; no single architecture dominates across problem classes. Physics-informed DeepONets~\citep{wang2021learning,mandl2025separable} replace data losses with residual losses but retain the fixed-parameter-box limitation, and transfer-learning approaches under conditional shift~\citep{weiss2016survey, goswami2022deep} extend a trained operator to new domains only with labeled target data at the shifted parameters. Our Part 2 reuses a frozen basis at an extrapolated parameter without any labeled data and without retraining the trunk.

\paragraph{PINNs and their failure modes.} Physics-informed neural networks~\citep{raissi2019physics} (alongside related residual-minimization schemes such as the deep Galerkin method~\citep{sirignano2018dgm}, and surveyed broadly in~\citep{karniadakis2021physics}) solve PDEs without labeled data but are notoriously hard to optimize. A substantial literature documents training pathologies in such coordinate-based and physics-informed networks: spectral bias~\citep{tancik2020fourier, wang2021eigenvector}, unbalanced loss gradients~\citep{wang2021understanding}, a neural-tangent-kernel account of why training stalls~\citep{wang2022and}, causality violations in time~\citep{wang2024respecting}, and instability at depth~\citep{Wang2024PirateNets}. Remedies span loss re-weighting and self-adaptive weights~\citep{wang2021understanding, wang2022and, mcclenny2023self}, residual-based adaptive sampling~\citep{wang2026causality,wu2023comprehensive}, adaptive activations~\citep{jagtap2020adaptive}, gradient-enhanced objectives~\citep{yu2022gradient}, architectural changes~\citep{Wang2024PirateNets}, and consolidated training recipes~\citep{wang2023expert}. Less studied are failures of basin selection: on problems whose residuals admit multiple physically distinct minima (bifurcating branches, symmetry-related states, trivial constant solutions) a from-scratch PINN can converge to a spurious solution branch that does not match the reference field, even when the residual is near zero. While temporal and adaptive sampling remedies address erroneous local minima in specific PDEs like the Allen--Cahn equation~\citep{wang2024respecting, wang2026causality}, our results address broader failures of basin selection that the other remedies above do not target.

\paragraph{Solving the lid-driven cavity.} The lid-driven cavity is the canonical incompressible-flow benchmark: multigrid finite-difference tables~\citep{ghia1982high}, spectral benchmarks~\citep{Botella1998Spectral}, high-Reynolds fine-grid studies~\citep{erturk2005numerical}, and an $8192\times8192$ Richardson-extrapolated solution~\citep{marchi2021lid} remain the standard yardstick. PINNs struggle far more: rising Reynolds number sharpens the corner singularities and destabilizes residual minimization. \citet{wang2023solution} show vanilla PINNs at $\Real=2000$--$5000$ admit two residual minimizers (one matching DNS, one unphysical) restored to uniqueness via an added entropy-viscosity term. State-of-the-art single-case accuracy comes from the residual-adaptive PirateNet architecture~\citep{Wang2024PirateNets}, our point of comparison at $\Real=3200$. These works solve one cavity at a time; we instead extrapolate a frozen operator basis, letting a crude teacher supply the physical basin.

\paragraph{Knowledge distillation.} Traditionally, knowledge distillation originates in model compression~\citep{hinton2015distilling}, where a smaller student network is trained to match a larger, highly accurate teacher. In this classical setup, the student's performance is inherently capped by a near-perfect teacher. Our approach differs fundamentally: the teacher is a neural operator providing only a crude out-of-distribution estimate, and the student is explicitly expected to surpass it through subsequent physics-based optimization. Accordingly, the distillation weight decays and ultimately vanishes rather than remaining active throughout training. This shares intuition with scientific computing approaches that use learned models as initial guesses, including warm-start strategies for PINNs~\citep{Wang2024PirateNets} and iterative solvers~\citep{eshaghi2026nows}. However, in contrast to methods that either apply the prior only at initialization~\citep{eshaghi2026nows} or rigidly freeze the prior representations~\citep{desai2021one, goswami2020transfer, chakraborty2021transfer}, our framework retains the teacher as a persistent but decaying constraint. This provides necessary guidance while the student explores the solution landscape, progressively relaxing the imperfect teacher's influence before transferring full control to the governing physics.

\section{Methodology}
\label{sec:method}

\subsection{Shared architecture}
\label{sec:architecture}

Let $s(\bx;\bp)$ denote the unknown solution field on domain $\Omega(\bp)$ for parameters $\bp\in\R^{d_p}$. We employ strong boundary enforcement by defining the ansatz as:
\begin{equation}
    \label{eq:ansatz}
    s(\bx;\bp) \;=\; g(\bx;\bp) \;+\; \mathcal{M}(\bx)\,\sum_{k=1}^K b_k(\bp)\,\varphi_k(\bx),
\end{equation}
where $\{\varphi_k\}_{k=1}^K$ is a spatial basis produced by a trunk network, $K$ is the number of trunk basis functions (equivalently, the number of branch outputs), and $b_k(\bp)$ are the branch coefficients, collected into the vector $\bb(\bp)=(b_1(\bp),\dots,b_K(\bp))$. Here, $g(\bx;\bp)$ is a known function satisfying the boundary conditions (the boundary lift), and $\mathcal{M}(\bx)$ is a distance multiplier that evaluates to zero on the boundary $\partial\Omega$. Enforcing boundary data exactly via this construction eliminates the need for boundary loss terms and stabilizes training~\citep{lu2021physics}.

For the Allen--Cahn problem, the lift $g$ also incorporates a non-zero initial profile. Because the trivial solution ($s=0$) satisfies the Allen--Cahn PDE exactly, this addition is required to prevent the network from collapsing into the non-physical zero branch. Consequently, no boundary loss term enters training, and the objective relies strictly on the mean-squared interior PDE residual, $\mathcal{R}$ (the problem-specific residual operator, given explicitly for each of the three PDEs in Appendix~\ref{app:per-problem}):
\begin{equation}
    \label{eq:base-loss}
    \mathcal{L}_{\mathrm{PDE}}(\bp) \;=\; \Big\|\mathcal{R}\big[s(\cdot;\bp)\big]\Big\|_{L^2(\Omega(\bp))}^2,
\end{equation}
averaged over parameter batches drawn from a bounded base box $\mathcal{B}\subset\R^{d_p}$. Throughout, we report a \mbox{relative-$L^2$} accuracy against a reference field $s_{\mathrm{ref}}$ (analytical or finite-difference, per problem):
\begin{equation}
    \label{eq:accuracy}
    \mathrm{Acc} \;=\; 1 - \varepsilon_{L^2}, \qquad
    \varepsilon_{L^2} \;=\; \frac{\|\hat{s} - s_{\mathrm{ref}}\|_{L^2}}{\|s_{\mathrm{ref}}\|_{L^2}},
\end{equation}
where $\hat{s}$ is the predicted field. Table~\ref{tab:main-results} reports $\varepsilon_{L^2}$ (labeled error) for the operator and both PINNs, alongside the absolute performance gain (labeled Improvement), defined as $\varepsilon_{L^2,\mathrm{baseline}} - \varepsilon_{L^2,\mathrm{distilled}}$. Everywhere else in the paper, performance is reported in terms of $\mathrm{Acc}$ (labeled accuracy).
All three problems in \S\ref{sec:experiments} instantiate the ansatz~\eqref{eq:ansatz} and train against the residual objective~\eqref{eq:base-loss}; they differ in three architectural aspects summarized in Table~\ref{tab:axes}: (i) the procedure used to construct the anchor coefficients after freezing the spatial basis; (ii) the order to which the multiplier $\mathcal{M}$ vanishes on the boundary; and (iii) the spatial bandwidth of the trunk features. Additional implementation-level differences, including residual discretization and a problem-specific regularizer, are specified in Appendix~\ref{app:differences}.

\begin{table}[htbp]
\centering
\small
\renewcommand{\arraystretch}{1.15}
\caption{Problem-specific design choices across the evaluated PDEs.}
\label{tab:axes}
\begin{tabularx}{\textwidth}{@{}p{0.20\linewidth} YYY@{}}
\toprule
 & Burgers & Allen--Cahn & Cavity \\
\midrule
Field / operator & scalar $u$; $u u_x-\nu u_{xx}=0$ & scalar $u$; $\lap u + \lambda(u{-}u^3){=}0$ & stream $\psi$; vorticity transport \\
Parameters $\bp$ & $(\nu, L)$ & $(\lambda, H)$ & $(\Real, H)$ \\
Order / dim. & 2nd-order, 1D & 2nd-order, 2D & 4th-order, 2D \\
Design choice (i): Part 1 anchors & self-supplied from branch & data-free Gauss--Newton on frozen basis & FD reference reprojected onto basis (8 solves) \\
Design choice (ii): boundary lift & 1st-order multiplier & seed lift $\ell_0$ + 1st-order mult. & 2nd-order multiplier + lid lift \\
Design choice (iii): trunk bandwidth & moderate, isotropic $\sigma=6$ & mild, isotropic ($\sigma{=}[3,3]$) & moderate, anisotropic $\sigma=[\sigma_\xi,\sigma_\eta]$ \\
Part 2 and 3 target & larger $L$ & $\lambda$ below base box (near onset) & larger $\Real$ (a geometry variant is reported in Appendix~\ref{app:extra}) \\
\bottomrule
\end{tabularx}
\end{table}

\subsection{Part 1: base operator}
\label{sec:part1}

Trunk and branch are trained jointly by minimizing~\eqref{eq:base-loss} over $\bp\sim\mathrm{Unif}(\mathcal{B})$. Upon convergence, the trunk is frozen, rendering $\{\varphi_k\}$ a fixed function space. An anchor set $\{c_i\}$ is then constructed at a small collection of base points $\{p_i\} \subset B$. These anchors act as trusted reference coefficients within the training domain, which will later be interpolated to form a structural prior for out-of-distribution extrapolation in Part 2. The three problems differ only in how these anchors are formed, exhibiting a spectrum of increasing reliance on external information (Table~\ref{tab:axes}, Design choice i). Burgers stores the trained branch output directly. Allen--Cahn refines it through a damped Gauss--Newton solve of the residual on the frozen basis, utilizing only the residual and its exact Jacobian (with no reference field). The cavity computes eight finite-difference reference fields at $\Real\in\{100,200,\dots,800\}$ with fixed $H{=}1$, reprojecting each onto the frozen basis via ridge least squares in the velocity metric. These eight solves represent the only external data the main cavity pipeline ever consumes.\footnote{The aspect-ratio extrapolation reported separately in Appendix~\ref{app:extra} trains an independent base operator over the full $(\Real,H)$ box with $12$ anchors on $\{100,200,300,400\}\times\{1.0,1.5,2.0\}$; see Table~\ref{tab:hyper} for both configurations.}

\subsection{Part 2: extrapolation beyond the base box}
\label{sec:part2}

While parameter continuation is common in PINN curricula to avoid failure modes~\citep{krishnapriyan2021characterizing}, unguided extrapolations often drift into non-physical configurations that spuriously minimize the residual. To prevent this, the base anchors are interpolated by a low-degree polynomial $\cvec_{\mathrm{cont}}(\bp)$ fit independently for each of the $K$ modes, acting as a continuation prior. To reach a target $\bp^\star\in\R^{d_p}\setminus\mathcal{B}$, the branch is retrained at $\bp^\star$ under two regularizing mechanisms that require no reference solution, keeping the extrapolation entirely data-free.

\paragraph{Bounded correction.} At the target, the branch network is re-optimized to output a raw coefficient correction $\Delta\bb(\bp)$ (analogous to $\bb(\bp)$ in \eqref{eq:ansatz}, but produced by a fresh Part-2 branch head rather than the Part-1 branch). Rather than using $\Delta\bb(\bp)$ directly as the coefficients, it is added to the continuation prior as a bounded perturbation,
\begin{equation}
    \label{eq:compose}
    \bb_{\mathrm{comp}}(\bp) \;=\; \cvec_{\mathrm{cont}}(\bp) \;+\; \alpha\,\tanh\!\big(\Delta\bb(\bp)\big),
\end{equation}
with $\alpha$ representing a hard cap on the departure from the prior. If the prior is correct, $\Delta\bb=0$ is optimal.

\paragraph{Gram-weighted continuation penalty.} Training minimizes
\begin{equation}
    \label{eq:part2loss}
    \mathcal{L}_{\mathrm{total}} \;=\; \mathcal{L}_{\mathrm{PDE}}(\bb_{\mathrm{comp}}) \;+\; \beta\,\big\|\bb_{\mathrm{comp}} - \cvec_{\mathrm{cont}}\big\|_{\Gram(H)}^2,
\end{equation}
where $\Gram(H)$ is the field-metric Gram matrix of the frozen basis, assembled on the physical geometry at aspect ratio $H$ via the isoparametric map of Appendix~\ref{app:per-problem} (for the 1D and fixed-geometry problems, $\Gram(H)$ reduces to a constant matrix $\Gram$). Penalizing coefficient deviations in the field metric (rather than the raw coefficient space) ensures the objective strictly penalizes modifications that alter the physical field. The initial scalar weight $\beta$ is calibrated once per target so that the penalty term contributes a fixed, predetermined fraction of the total initial objective loss. From that calibrated value, $\beta$ is either held constant for the remainder of the step or annealed exponentially toward a small non-zero floor over the target's training budget, letting the PDE residual increasingly govern convergence as training proceeds; the problem-specific choices are given in Appendix~\ref{app:hyperparams}.

The fundamental extrapolation operation projects the base anchors to a new target parameter. If the target $\bp^\star$ is sufficiently close to the training domain, this is executed in a single shot: the prior is fit once on the Part 1 anchors, and the branch is retrained directly at $\bp^\star$ using~\eqref{eq:compose} and \eqref{eq:part2loss}. However, when a single step spans too large a parameter distance, the extrapolation employs an incremental self-supply strategy. In this multi-stage approach, the model extrapolates to intermediate targets; upon convergence, these intermediate solutions are appended to the anchor pool, and the prior is refitted before advancing to the next stage. For example, the principal cavity $\Real$-extrapolation utilizes a four-stage continuation (with a 1D prior over $1/\Real$), the geometry variant fits a 2D prior over $(1/\Real, H)$ (Appendix~\ref{app:per-problem}), and Allen--Cahn requires a two-stage sequence (Appendix~\ref{app:differences}).

The goal of this stage is not high target accuracy, but to generate a coefficient field sufficient to initialize Part 3, which then refines the solution. The Allen–Cahn problem illustrates this: near the pitchfork bifurcation, the true branch amplitude scales as $\sqrt{\lambda-\lambda_1}$ (with $\lambda_1$ the critical rate), a form the analytic polynomial prior cannot represent. As a result, the extrapolated field degrades predictably at the near-onset target.

\subsection{Part 3: Guarded Distillation into a Fresh Solver}
\label{sec:part3}

\begin{figure}[htbp]
    \centering
    \includegraphics[width=0.9\textwidth]{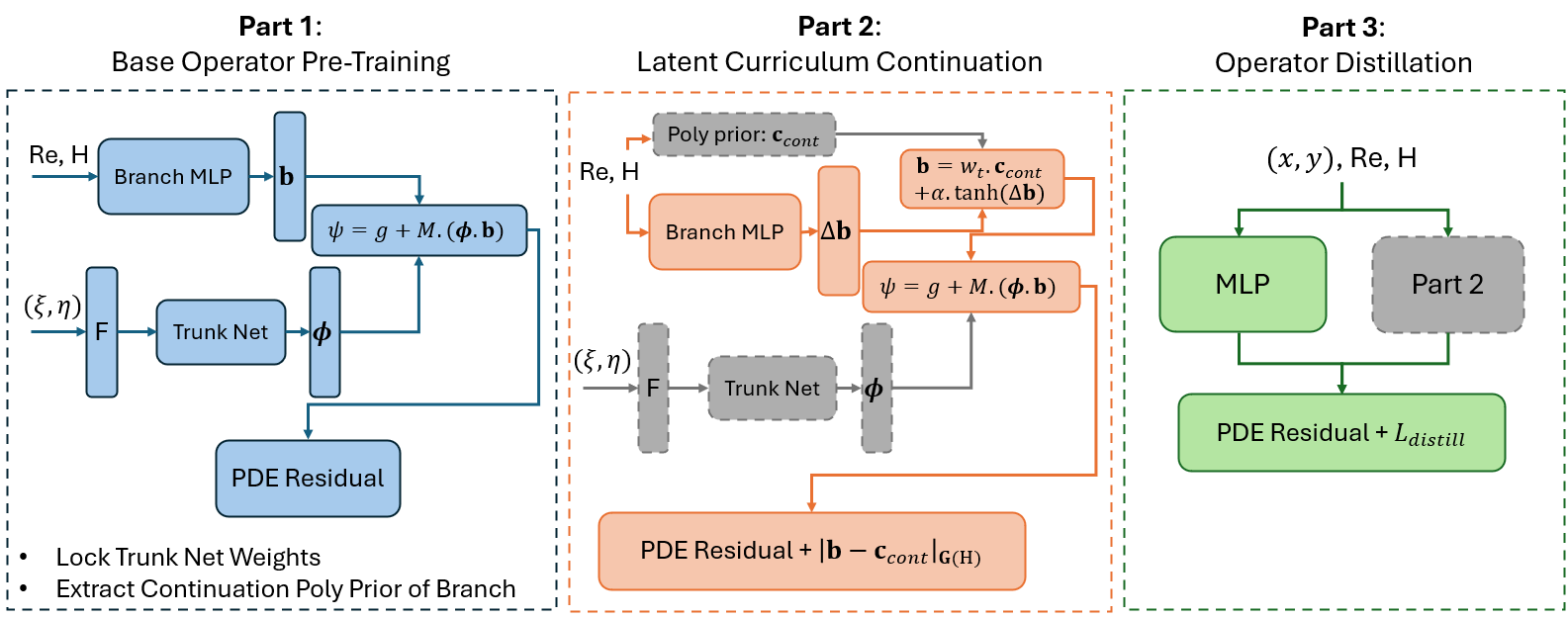}
    \caption{Three-stage framework schematic.}
    \label{fig:schematic}
\end{figure}

At the target $\bp^\star$, the composited field from~\eqref{eq:compose} is passed to a fresh MLP-PINN $u_\phi$ trained from scratch on the physics residual. Crucially, this student network abandons the trunk-branch decomposition and the specialized boundary lift utilized in Part 1. The training loss over optimization step $t$ is defined as:
\begin{equation}
    \label{eq:distil-loss}
    \mathcal{L}(\phi;t) \;=\; \big\|\mathcal{R}[u_\phi]\big\|_{L^2(\Omega)}^2 +\lambda_{\mathrm{BC}}\mathcal{L}_{\mathrm{BC}} +w(t)\big\|u_\phi-u_{\mathrm{teacher}}\big\|_{\Gamma(\bx)}^2,
\end{equation}
where $\mathcal{L}_{\mathrm{BC}}$ penalizes boundary-condition violations, $u_{\mathrm{teacher}}$ is the Part~2 prediction, $w(t)$ is the distillation schedule, and $\Gamma(\bx)$ is an optional spatial reliability gate (Appendix~\ref{app:distill}). To demonstrate modularity, we employ two strategies. Allen--Cahn and the primary Re-extrapolated cavity use an ungated ($\Gamma \equiv 1$) piecewise-linear $w(t)$ that vanishes during the final training phase. Conversely, the Burgers and geometry-extrapolated cavity configurations utilize a physics-adaptive gate: $w(t)$ tracks the student's residual, while $\Gamma(\bx)$ locally attenuates the teacher's influence at high-residual points to tolerate localized teacher inaccuracies. Both schedules strictly vanish before training concludes, leaving the PDE and boundary constraints to govern final convergence. Figure~\ref{fig:schematic} summarizes the complete three-stage framework.

\section{Experiments}
\label{sec:experiments}

We evaluate the framework through a progressive sequence of experiments designed to address four core objectives. First, we use the viscous Burgers equation to establish whether an approximate, out-of-distribution teacher can accelerate a PINN (\S\ref{sec:results-burgers}). Second, we analyze the Allen--Cahn equation to determine why this transfer succeeds in basin selection where pure residual minimization fails (\S\ref{sec:results-ac}). Third, we evaluate the lid-driven cavity to test if the mechanism scales to complex, high-dimensional systems far outside their training distribution (\S\ref{sec:results-cavity}). Finally, an ablation study isolates precisely which component of the framework drives this success (\S\ref{sec:ablation-part2}).

\begin{table}[t]
\centering
\small
\caption{Performance summary across the three evaluated PDEs.}
\label{tab:main-results}
\begin{tabular}{lccc}
\toprule
 & Burgers & Allen--Cahn & Cavity \\
\midrule
Training range
& $(\nu,L)\in[0.10,0.20]\times[1,2]$
& $(\lambda,H)\in[35,80]\times[1,2]$
& $(\mathrm{Re},H)\in[100,800]\times\{1\}$ \\

Target
& $(\nu,L)=(0.05,4)$
& $(\lambda,H)=(12.5,3)$
& $(\mathrm{Re},H)=(3200,1)$ \\

OOD$^*$ distance
& $2\times$
& $0.35 \times$ lower box edge
& $4\times$ \\

\midrule
Operator: in-distribution error
& $0.003 \pm 0.002$
& $0.0015 \pm 0.0004$
& $0.2586 \pm 0.1294$ \\

Operator: OOD error
& $0.12$
& $0.78$
& $0.53$ \\

Baseline PINN error
& $0.4123 \pm 0.1601$
& $1.00 \pm 0.00$
& $0.9800 \pm 0.0248$ \\

Distilled PINN error (\textbf{Ours})
& $0.0209 \pm 0.0043$
& $0.0037 \pm 0.001$
& $0.0663 \pm 0.0099$ \\

\midrule
Improvement
& $0.3914 \pm 0.1635$
& $0.9963 \pm 0.001$
& $0.9137 \pm 0.0341$ \\

\bottomrule
\end{tabular}

\vspace{4pt}
{\raggedright \footnotesize $^*$OOD stands for out of distribution.\par}
\end{table}

\subsection{Experiment 1 (Basic transfer): Viscous Burgers (1D)}
\label{sec:results-burgers}
\begin{figure}[htbp]
    \centering
    \includegraphics[width=0.75\textwidth]{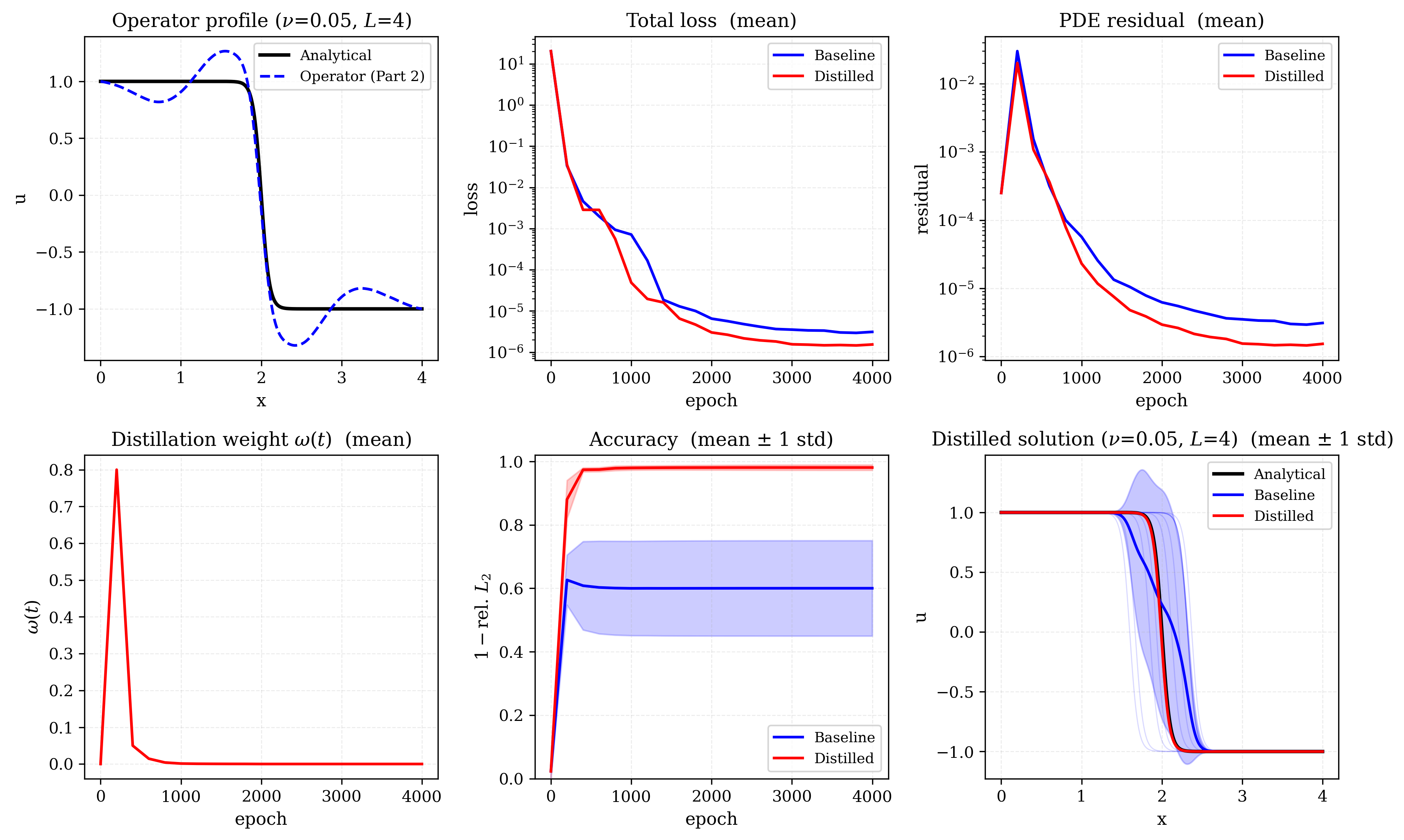}
    \caption{Burgers equation training metrics aggregated over $n=5$ seeds. Panel (a) illustrates the Part 2 operator extrapolation versus the analytical reference. The remaining five panels detail the Part 3 distilled PINN performance.}
    \label{fig:burgers_results}
\end{figure}

The 1D viscous Burgers equation establishes the fundamental transfer mechanism in a controlled setting featuring a unique monotone solution manifold, one internal transition layer, and a Cole--Hopf closed form for exact evaluation. The base operator is trained on $L\in[1,2]$ in Part 1 and extrapolated to $L=4$ at $\nu=0.05$ in Part 2, extending twice beyond the maximum training length. Figure~\ref{fig:burgers_results}(a) illustrates the frozen-trunk operator at this out-of-distribution target: it captures the correct shock location and outer branches but exhibits an overshoot at the transition layer. This represents the precise approximate-yet-structurally-accurate prior the framework is designed to exploit. Across $n=5$ independent seeds, the distilled student removes the overshoot and matches the analytical solution to graphical accuracy in every run (Figure~\ref{fig:burgers_results}(f)). It reaches $97.91\% \pm 0.43\%$ mean accuracy versus $58.77\% \pm 16.01\%$ for the from-scratch baseline, an improvement of $39.14\% \pm 16.35\%$ that holds in all 5/5 seeds. Beyond the higher mean, distillation markedly reduces run-to-run variance: the baseline's final accuracy is highly seed-dependent and occasionally collapses to a poor local solution, while the distilled student consistently converges to a tight neighborhood of the true solution regardless of initialization. Distillation also converges faster and to a lower final residual, with the reliability gate closing as the student's residual falls (Figure~\ref{fig:burgers_results} d).

\subsection{Experiment 2 (Basin selection): Steady Allen--Cahn (2D)}
\label{sec:results-ac}
\begin{figure}[htbp]
\centering
\includegraphics[width=0.75\textwidth]{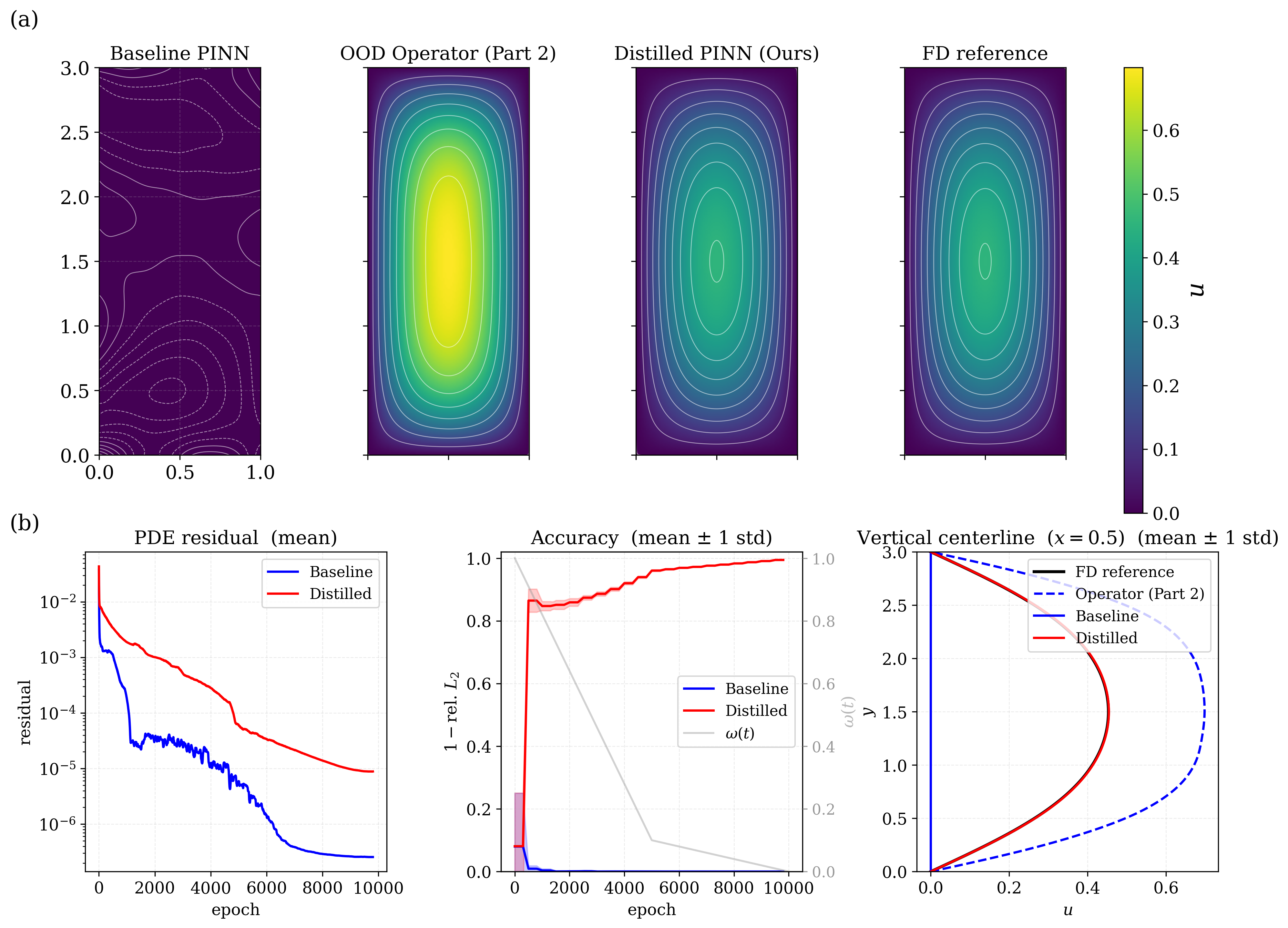}
\caption{Allen--Cahn training metrics ($\lambda=12.5$, $H=3$), aggregated over $n=5$ seeds, versus finite difference reference.}
\label{fig:ac}
\end{figure}

This experiment demonstrates how the teacher resolves a solution-branch ambiguity that pure residual minimization cannot. At $\lambda$ close to the pitchfork bifurcation $\lambda_1(H) = \pi^2(1+1/H^2)$, the principal positive branch $u^\star>0$ exhibits an $O(\sqrt{\lambda-\lambda_1})$ amplitude, while the trivial solution ($u\equiv0$) satisfies the residual exactly. We evaluate the framework at $\lambda=12.5$ and $H=3$ (where $\lambda_1(3)\approx10.97$). This target resides approximately $14\%$ above the onset, with a true peak amplitude of $\approx 0.45$.
Figure~\ref{fig:ac} (a) illustrates this dynamic. Baseline PINN collapses to the flat $u\equiv0$ field, identifying a mathematically correct but physically trivial global minimizer; across $n=5$ seeds, this collapse is total and deterministic, with $0.00\% \pm 0.00\%$ accuracy and a peak amplitude of only $0.0002\pm0.0002$ against a true peak of $0.4530$. The Part 2 operator (second panel) correctly identifies the single-hump structure but over-predicts the amplitude, as the polynomial prior cannot precisely capture the $\sqrt{\lambda-\lambda_1}$ onset behavior, giving it only $32\%$ accuracy on its own. Distilling this operator provides the necessary prior to pull the student into the correct basin of attraction despite the teacher's own imprecision. Subsequent residual minimization sharpens the field: Distilled PINN matches the finite-difference reference in both shape and amplitude, reaching $99.63\% \pm 0.10\%$ mean accuracy across seeds. Figure~\ref{fig:ac} (b) confirms this quantitatively: accuracy climbs as the distillation weight anneals to zero, and the centerline profile aligns with the reference. This confirms the central hypothesis of the framework: an imperfect operator is sufficient because its role is solely to identify the correct physical basin, not to provide the exact solution. Here, a teacher accurate to only $32\%$ still reliably rescues every seed from the trivial minimizer.

\subsection{Experiment 3 (Large OOD extrapolation): Lid-driven cavity (2D)}
\label{sec:results-cavity}
To demonstrate scalability, the lid-driven cavity introduces a fourth-order, two-dimensional system extrapolated substantially outside its base box. The primary target is set to $\Real=3200$ (far outside the $\Real\in[100,800]$ training domain) to match the state-of-the-art single-case PINN benchmark established by~\citet{Wang2024PirateNets}. A secondary experiment, extrapolating aspect ratio rather than Reynolds number, is reported in Appendix~\ref{app:extra}.

\begin{figure}[!htbp]
  \centering
  \includegraphics[width=0.75\linewidth]{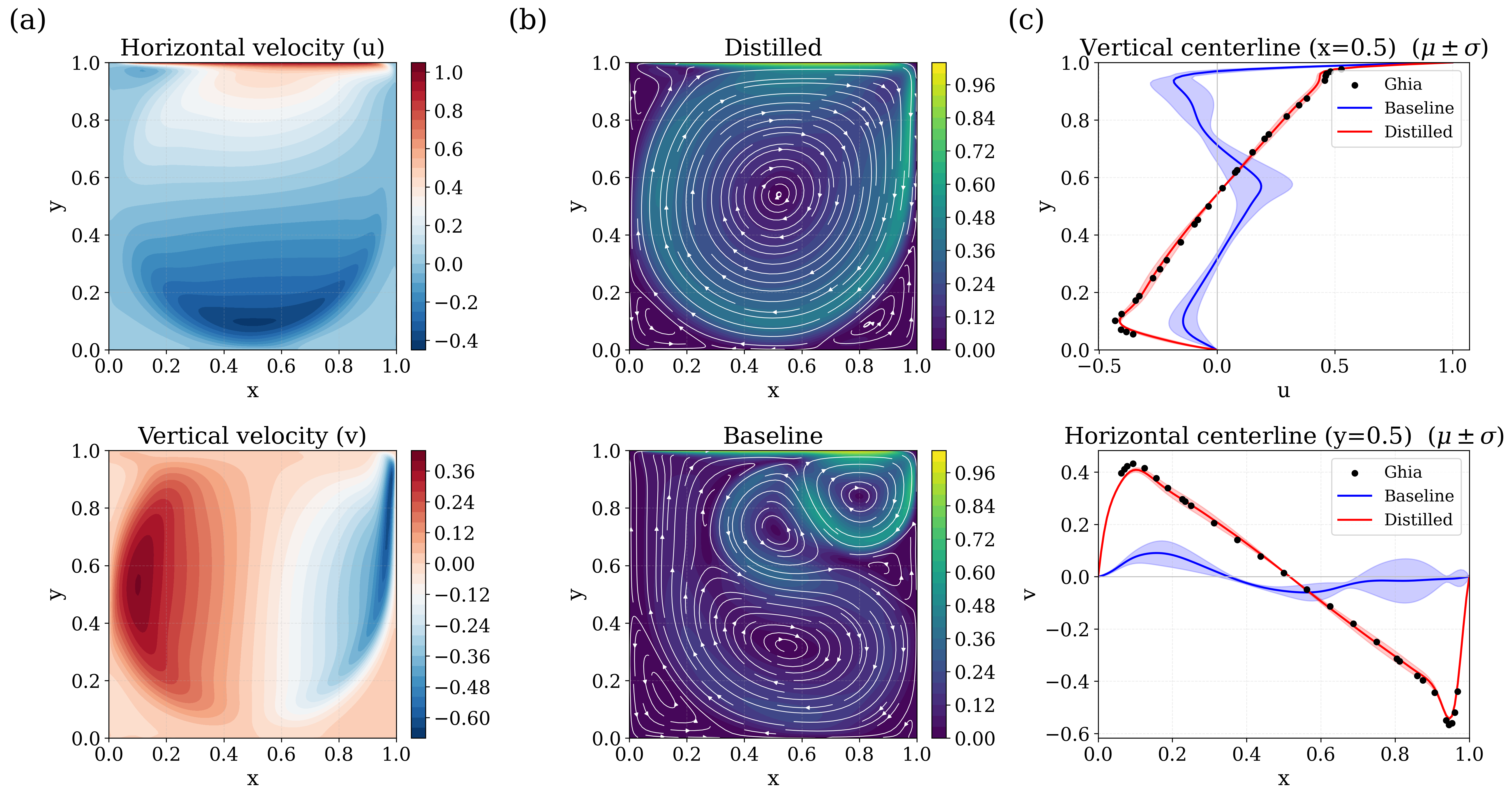}
  \caption{Lid-driven cavity at $\mathrm{Re}=3200$. (a) Predicted $u$ and $v$ fields. (b) Velocity magnitude and streamlines for the distilled and baseline students. (c) Centerline profiles at $x=0.5$ and $y=0.5$ against~\cite{ghia1982high}. Fields and streamlines show the ensemble mean across $n=5$ seeds.}
  \label{fig:ns_panel}
\end{figure}

Figure~\ref{fig:ns_panel} illustrates the distilled student at $\Real=3200$. Aggregated across $n=5$ independent seeds, the $u$ and $v$ fields (a) successfully capture the primary vortex and thin wall jets. The streamlines (b) reproduce the primary recirculation and the two lower corner vortices in the distilled model, features that the baseline completely fails to capture. Consequently, the centerlines (c) tightly align with the benchmark points established by~\cite{ghia1982high} and \cite{marchi2021lid}, whereas the baseline deviates severely.

\begin{figure}[!htbp]
  \centering
  \includegraphics[width=0.75\linewidth]{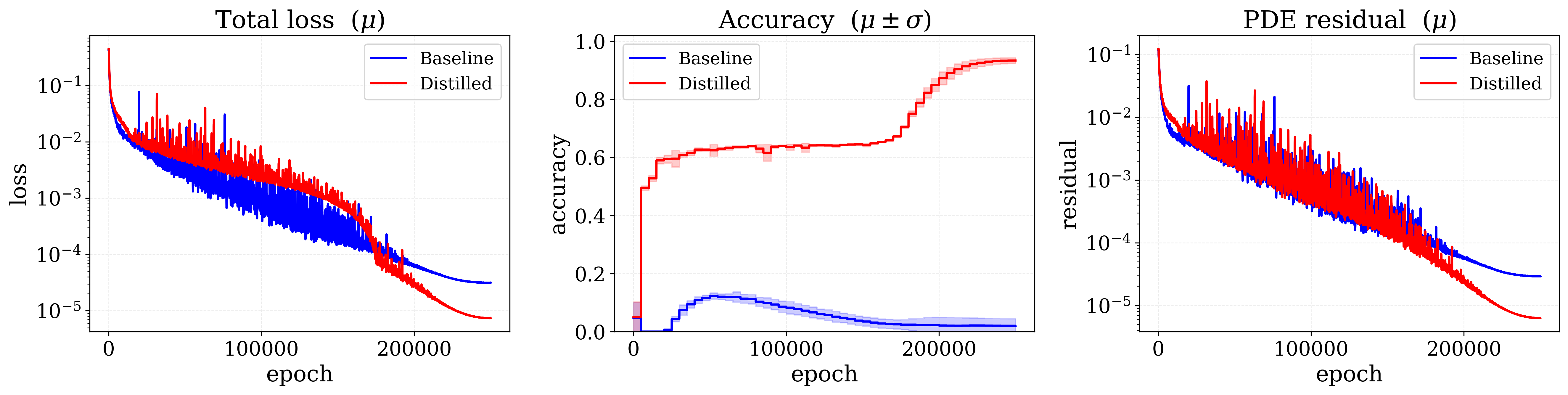}
  \caption{Lid-driven cavity ($\mathrm{Re}=3200$) training metrics. Solid lines and shaded regions represent $\mu \pm \sigma$ across $n=5$ independent seeds.}
  \label{fig:ns_curves}
\end{figure}

Figure~\ref{fig:ns_curves} details the optimization trajectory: the baseline stalls at near-zero accuracy with a noisy residual, while the distilled student steadily climbs to high accuracy at a lower final residual, utilizing the identical architecture and step budget. Quantitatively, the baseline PINN entirely fails to converge to the correct physical state, yielding a mean accuracy of just $2.00\% \pm 2.48\%$. In contrast, the distilled student consistently resolves the solution-branch ambiguity, achieving a mean accuracy of $93.37\% \pm 0.99\%$. This represents a robust accuracy improvement of $91.4\% \pm 3.4\%$, an acceleration that holds consistently across all $5/5$ initialized seeds. Beyond the higher mean performance, distillation markedly reduces run-to-run variance, confirming that the operator initialization effectively guards against the inherent sensitivity of PINNs to random parameter initialization.

For computational context, Table~\ref{tab:piratenet-comparison} compares the Part~3 student with~\citet{Wang2024PirateNets}: our student uses roughly $20\%$ of the parameters and $40\%$ of the optimization steps ($250$K vs.\ $630$K), but more point evaluations. End-to-end, Parts~1--3 require $546$K optimization steps, but note that Part~1 is a one-time base-box cost reused across subsequent targets.

\begin{table}[htbp]
\centering
\caption{Part~3 computational comparison at $\mathrm{Re}=3200$ with~\citet{Wang2024PirateNets}.}
\label{tab:piratenet-comparison}
\begin{tabular}{lcc}
\toprule
                                    & Distilled PINN (\textbf{Ours}) & PirateNet \\
\midrule
Rel.\ $L^2$ error against~\cite{ghia1982high}           & $0.0663 \pm 0.0099$            & $0.0421$             \\
Depth $\times$ width                & $5 \times 256$                 & $18 \times 256$      \\
Trainable parameters                & 297\,K                         & $\sim$1.32\,M        \\
Total training steps                & 250\,K                         & 630\,K               \\
Total point-evaluations             & $6.5 \times 10^{9}$            & $2.6 \times 10^{9}$  \\
\bottomrule
\end{tabular}
\end{table}

\subsection{Experiment 4 (Mechanistic ablation): The role of Part 2}
\label{sec:ablation-part2}

Section~\ref{sec:results-cavity} presents the full framework, where Part 2 extrapolates the frozen-trunk operator over a four-stage $\mathrm{Re}$ sequence (\S\ref{sec:part2}) before distillation. This raises a key ablation: does incremental continuation uniquely determine basin selection, or is a direct evaluation of the Part 1 operator at $\mathrm{Re}=3200$ an adequate prior? To test this, we keep the Part 3 architecture, the piecewise-linear distillation schedule, and the $250$K-step training budget fixed, but initialize the teacher coefficients by directly evaluating the Part 1 branch at the target:
$c = b_{\mathrm{Part~1}}(\mathrm{Re}{=}3200, H{=}1)$.
This extrapolates the neural operator fourfold beyond its training range ($\mathrm{Re}\in[100,800]$) without the regularizing polynomial continuation prior~\eqref{eq:compose}. We then compare this ablated model with the full framework and an unguided baseline PINN.

Table~\ref{tab:part2-ablation} and Figure~\ref{fig:cavity-ablation} summarize the ensemble results across $n=5$ independent seeds. Relying exclusively on the unguided Part 1 teacher consistently fails to identify the correct physical basin. As depicted in the mean velocity fields, the student's primary vortex is mislocated and the lower corner vortices are entirely absent, resulting in severe deviations from the benchmark~\citep{ghia1982high}. Quantitatively, the ablated model yields a highly variable mean accuracy of $27.09\% \pm 13.49\%$. While this slightly outperforms the catastrophic collapse of the baseline PINN ($2.00\% \pm 2.48\%$), it falls drastically short of the full framework's $93.37\% \pm 0.99\%$.

\begin{table}[htbp]
\centering
\small
\caption{Ablation of the Part~2 continuation stage for the cavity at $\mathrm{Re}=3200$, $H=1$.}
\label{tab:part2-ablation}
\begin{tabular}{lccc}
\toprule
 & Full framework & No Part 2 & Baseline PINN \\
\midrule
Teacher
& Continued operator
& Direct Part-1 operator 
& None \\

Relative $L^2$ accuracy
& $93.37\% \pm 0.99\%$
& $27.09\% \pm 13.49\%$
& $2.00\% \pm 2.48\%$ \\

PDE residual
& $6.3\times10^{-6}$
& $1.3\times10^{-5}$ 
& $2.9\times10^{-5}$ \\

\bottomrule
\end{tabular}
\end{table}

Crucially, this represents a basin-selection failure rather than an optimization failure. The mean PDE residual for the ablated student converges cleanly to $1.3\times10^{-5}$, which is strictly comparable to both the full framework's $6.3\times10^{-6}$ and the baseline PINN's $2.9\times10^{-5}$ (Table~\ref{tab:part2-ablation}). This demonstrates that all three networks effectively minimize the governing physical equations, yet both the unguided baseline and the ablated model converge to steady states that deviate substantially from the~\citet{ghia1982high} reference. This behavior perfectly characterizes the basin-selection vulnerability the proposed framework is designed to circumvent (\S\ref{sec:results-ac}).

\section{Conclusion}
We introduced a three-stage framework that decouples the distinct failure modes of operator learning and physics-informed neural networks. The central finding of this work, demonstrated across three nonlinear PDEs, is that an approximate neural operator need not be highly accurate to be effective; it needs only to initialize a student network within the correct basin of attraction, after which residual minimization governs final optimization. Our analyses confirm this mechanism, demonstrating that distilled students robustly recover the true physics in scenarios where standard PINNs deterministically collapse or stall at non-physical steady states. Future directions include extending the continuation prior to better capture bifurcation onsets, evaluating the framework in time-dependent multi-physics environments, and integrating this distillation strategy with state-of-the-art architectures~\citep{Wang2024PirateNets} to yield compounding improvements in optimization stability.

\subsubsection*{Reproducibility statement}
All experimental hyperparameters are given in Appendix~\ref{app:hyperparams}. The finite-difference reference solvers used for scoring and, for the cavity problem, anchor formation are standard, and their configurations are specified in Appendix~\ref{app:fd}. An anonymized implementation and pretrained models are available at \url{https://anonymous.4open.science/r/PINN-Distillation-prototype-D3AF/}.

\subsubsection*{AI use statement}
During the preparation of this work, the authors used Claude (Anthropic) for assistance in writing the code used for analysis and evaluation, and to draft and revise manuscript text; Gemini (Google) was used to help improve manuscript text. The authors did not use generative AI tools for research ideation or experimental design; the three-stage framework and the extrapolation experiments described in this paper reflect the authors' own conceptual contributions. The authors also did not use generative AI tools for data analysis or interpretation of results. All AI-assisted work was reviewed before inclusion: LLM-generated code was verified and tested for correctness by the authors, and AI-drafted and AI-revised manuscript text was reviewed and edited by all the authors for accuracy and clarity. The authors take full responsibility for the content of the published article.

\appendix
\section{Per-problem instantiations of the shared architecture}
\label{app:per-problem}

\subsection{Domains and the isoparametric map}
The two-dimensional problems are posed on $\Omega_H = (0,1)\times(0,H)$; the one-dimensional problem on $(0,L)$. A reference domain $(\xi,\eta)\in[0,1]^2$ is mapped to physical coordinates by $(x,y)=(\xi,H\eta)$, yielding $\partial_x=\partial_\xi$ and $\partial_y=H^{-1}\partial_\eta$. Analogously, the one-dimensional reference coordinate $\xi\in[0,1]$ is mapped to the physical domain by $x=L\xi$, yielding $\partial_x = L^{-1}\partial_\xi$. The trunk basis is defined exclusively on the reference domain, such that geometry enters solely through algebraic chain-rule weights. Reference-space partial derivatives of the frozen basis remain geometry-independent; consequently, assembling the physical operator at a new geometry (i.e.\ a new $L$ or $H$) reduces to algebraic scaling.

\subsection{Burgers: PDE, ansatz, residual}
\label{app:burgers}
On $(0,L)$ with $u(0)=+A$, $u(L)=-A$ (we use $A=1$ in all experiments),
\begin{equation}
    u\,\partial_x u - \nu\,\partial_{xx} u = 0,
\end{equation}
with ansatz $u(\xi;\bp) = A(1-2\xi) + 4\xi(1-\xi)\sum_k b_k(\bp)\varphi_k(\xi)$ and residual $r = L^{-1} u\partial_\xi u - \nu L^{-2}\partial_{\xi\xi}u$. In the notation of~\eqref{eq:ansatz}, the boundary lift is $g(\xi;\bp) = A(1-2\xi)$ and the multiplier is $\mathcal{M}(\xi) = 4\xi(1-\xi)$, which vanishes to first order at $\xi\in\{0,1\}$ (Table~\ref{tab:axes}, Design choice ii).

\subsection{Allen--Cahn: PDE, ansatz, residual, bifurcation}
\label{app:ac}
On $\Omega_H$ with $u=0$ on $\partial\Omega_H$,
\begin{equation}
    \label{eq:ac-pde}
    \lap u + \lambda(u - u^3) = 0.
\end{equation}
The critical rate is $\lambda_1(H) = \pi^2(1+1/H^2)$, representing the first Dirichlet eigenvalue of $-\lap$ on the domain. Below $\lambda_1(H)$, the only solution is the trivial state $u\equiv 0$; above it, a pair $\pm u^\star$ bifurcates along the principal (nodeless, sign-definite) eigendirection of $-\lap$, scaling as $\|u^\star\|_\infty \sim \sqrt{\lambda-\lambda_1}$ near the onset and saturating to $\|u^\star\|_\infty \to 1$ with boundary layers of width $\sim 1/\sqrt{\lambda}$ further above the critical rate. Because the seed lift $\ell_0>0$ (defined below) biases the ansatz toward the positive member of this pair, we take $u^\star>0$, the principal positive branch, as the target solution throughout. The base training domain $\lambda\in[35,80], H\in[1,2]$ is restricted to the comfortably supercritical regime.

The ansatz employs a fixed non-trivial seed lift $\ell_0(\xi,\eta) = A_0\sin(\pi\xi)\sin(\pi\eta)$, with $A_0=0.3$ used in all experiments:
\begin{equation}
    u(\xi,\eta;\bp) = A_0 \sin(\pi\xi)\sin(\pi\eta) + 16\xi(1-\xi)\eta(1-\eta)\sum_k b_k(\bp)\varphi_k(\xi,\eta).
\end{equation}
In the notation of~\eqref{eq:ansatz}, $g(\xi,\eta;\bp) = \ell_0(\xi,\eta)$ (constant in $\bp$) and $\mathcal{M}(\xi,\eta) = 16\xi(1-\xi)\eta(1-\eta)$, which vanishes to first order on $\partial\Omega_H$ (Table~\ref{tab:axes}, Design choice ii). This formulation explicitly breaks the $u\equiv 0$ degeneracy at initialization; setting $b\equiv 0$ yields a positive interior profile rather than a trivial zero field. The seed function vanishes on $\partial\Omega_H$ and satisfies $\lap \ell_0 = -\pi^2(1+H^{-2})\ell_0 = -\lambda_1(H)\,\ell_0$: it is an eigenfunction of the linear Laplacian, with eigenvalue exactly equal to the critical bifurcation rate, but it is not an eigenfunction of the nonlinear residual~\eqref{eq:ac-pde}, so it is never itself a spurious fixed point of training.

\subsection{Cavity: PDE, ansatz, residual}
\label{app:cavity}
The streamfunction $\psi$ satisfies the vorticity-transport equation:
\begin{equation}
    -u\,\partial_x(\lap\psi) - v\,\partial_y(\lap\psi) + \Real^{-1}\bib\psi = 0,\quad u=\partial_y\psi,\ v=-\partial_x\psi,
\end{equation}
enforcing velocity Dirichlet conditions on all walls alongside a prescribed lid velocity $u_{\mathrm{lid}}$. The ansatz $\psi = g(\bx) + \mathcal{M}(\bx)\sum_k b_k\varphi_k(\bx)$ employs a lift function $g(\xi,\eta) = H u_{\mathrm{lid}}(\xi)(\eta^3-\eta^2)$ and a multiplier $\mathcal{M} = (16\xi(1-\xi)\eta(1-\eta))^2$. Because $\mathcal{M}$ vanishes to the second order on all boundaries (Table~\ref{tab:axes}, Design choice ii), both $\psi$ and its normal derivative $\partial_n\psi$ exactly satisfy the boundary conditions independent of the network coefficients.

\paragraph{Regularized lid profile.} A uniformly driven lid ($u=1$) introduces a velocity discontinuity at the two top corners where the moving boundary intersects the stationary side walls. This corner singularity generates unbounded vorticity, which frequently destabilizes high-$\Real$ residual training~\citep{wang2023solution}. To circumvent this, we regularize the boundary condition. The lid speed is maintained at exactly $1$ over the interior of the top wall and transitions smoothly to $0$ at each corner across a margin of width $\delta=0.12$:
\begin{equation}
    \label{eq:cavity-lid-ref}
    u_{\mathrm{lid}}(\xi)
    = S\!\Big(\tfrac{\xi}{\delta}\Big)\,
      S\!\Big(\tfrac{1-\xi}{\delta}\Big),
    \qquad
    \begin{aligned}
    S(t) &= 0, && t<0,\\
    S(t) &= 6t^5-15t^4+10t^3, && 0\le t\le 1,\\
    S(t) &= 1, && t>1.
    \end{aligned}
\end{equation}
where $S$ is Perlin's quintic smootherstep function \citep{perlin2002improving}. This formulation ensures matching values, slopes, and curvatures ($S(0){=}0$, $S(1){=}1$, $S'{=}S''{=}0$ at both domain ends), guaranteeing that the transition into the flat core introduces no gradient discontinuities up to the second derivative. With $\delta=0.12$, the lid remains exactly $1$ over the central $76\%$ of the span, modifying the physical driving force only within the narrow corner margins. This identical $u_{\mathrm{lid}}$ profile is applied to the lift $g$, the finite-difference reference solver, and the Part 3 student's boundary loss, ensuring the teacher and student solve the exact same regularized boundary-value problem.

\paragraph{Part 3 primitive-variable formulation.} Unlike Parts 1--2, which operate on the streamfunction $\psi$, the Part 3 cavity student is a primitive-variable PINN $u_\phi(\bx)=(u_\phi,v_\phi,p_\phi)$ trained directly on the steady incompressible Navier--Stokes equations:
\begin{align}
    u_\phi\,\partial_x u_\phi + v_\phi\,\partial_y u_\phi &= -\partial_x p_\phi + \Real^{-1}\big(\partial_{xx}u_\phi+\partial_{yy}u_\phi\big), \label{eq:momx}\\
    u_\phi\,\partial_x v_\phi + v_\phi\,\partial_y v_\phi &= -\partial_y p_\phi + \Real^{-1}\big(\partial_{xx}v_\phi+\partial_{yy}v_\phi\big), \label{eq:momy}\\
    \partial_x u_\phi + \partial_y v_\phi &= 0. \label{eq:cont}
\end{align}
Dirichlet velocity conditions are enforced softly: $u_\phi=v_\phi=0$ on the left, right, and bottom walls, and $u_\phi=u_{\mathrm{lid}}(x)$, $v_\phi=0$ on the top wall, using the identical regularized lid profile~\eqref{eq:cavity-lid-ref} applied throughout the pipeline. Because $p_\phi$ is determined only up to an additive constant, we impose a soft gauge penalty pinning its value at the domain center to zero, $p_\phi(0.5, H/2) \approx 0$. The complete Part 3 objective is
\begin{equation}
    \label{eq:distil-loss-cavity}
    \mathcal{L}(\phi;t) \;=\; \mathcal{L}_{\mathrm{mom},x} + \mathcal{L}_{\mathrm{mom},y} + \mathcal{L}_{\mathrm{cont}} \;+\; \lambda_{\mathrm{BC}}\,\mathcal{L}_{\mathrm{BC}} \;+\; \lambda_{\mathrm{gauge}}\,p_\phi(0.5,H/2)^2 \;+\; w(t)\,\mathcal{L}_{\mathrm{distill}}(t),
\end{equation}
where $\mathcal{L}_{\mathrm{mom},x}$, $\mathcal{L}_{\mathrm{mom},y}$, and $\mathcal{L}_{\mathrm{cont}}$ are the mean-squared residuals of~\eqref{eq:momx}--\eqref{eq:cont} over interior collocation points, $\mathcal{L}_{\mathrm{BC}}$ is the mean-squared boundary-velocity violation, and $\lambda_{\mathrm{BC}}$, $\lambda_{\mathrm{gauge}}$ are the momentum-normalized BC and gauge weights reported in Table~\ref{tab:hyper} (10.0 and 10.0 respectively for the main $\Real$-extrapolation target; the momentum terms carry unit weight). The distillation term is applied only to velocity:
\begin{equation}
    \label{eq:distill-term-cavity}
    \mathcal{L}_{\mathrm{distill}}(t) \;=\; \big\|u_\phi - u_{\mathrm{teacher}}\big\|_{\Gamma(\bx)}^2 \;+\; \big\|v_\phi - v_{\mathrm{teacher}}\big\|_{\Gamma(\bx)}^2,
\end{equation}
where $(u_{\mathrm{teacher}},v_{\mathrm{teacher}}) = (\partial_y\psi_{\mathrm{teacher}}, -\partial_x\psi_{\mathrm{teacher}})$ are the induced velocities of the Part~2 streamfunction field, obtained by automatic differentiation of the frozen Part~2 network. No distillation target is imposed on $p_\phi$: the streamfunction teacher carries no pressure information, so the pressure field is constrained solely by the momentum residuals and the gauge penalty. This primitive-variable formulation for Part 3 is used identically for both the $\Real$-extrapolation (\S\ref{sec:results-cavity}) and $H$-extrapolation (Appendix~\ref{app:extra}) cavity experiments; the Burgers and Allen--Cahn students, by contrast, retain the scalar-field loss~\eqref{eq:distil-loss} directly on $u_\phi$, since those problems have no pressure or continuity constraint.

Interior collocation points for the momentum and continuity residuals are not sampled uniformly: the wall-layer thickness scales as $\delta_{\mathrm{wall}}\sim\Real^{-1/2}$, comparable to or smaller than the spacing of a uniform sample, which otherwise lets the student smooth away the near-wall velocity extrema. We instead draw $40\%$ of interior points from thin near-wall bands (width $4\Real^{-1/2}$, biased toward each wall) and the remaining $60\%$ uniformly, applied identically to the baseline and distilled arms so it does not bias their comparison. Distillation query points are sampled uniformly rather than wall-clustered, since the teacher's velocity error is largest near the walls and clustering there would concentrate the distillation pull on its least reliable signal.

\subsection{Spatial bandwidth and spectral bias}
\label{app:bandwidth}
Standard coordinate-based networks exhibit spectral bias, so a random Fourier embedding is used to shift the burden of representing high-frequency features onto the analytical derivatives of the basis sinusoids. The required bandwidth scales with the PDE's order: a frequency-$k$ component entering an $n$-th-order residual is amplified by a factor of $\sim k^n$, bounded by $k^2$ for the second-order Burgers and Allen--Cahn systems but reaching $k^4$ for the fourth-order cavity system, further exacerbated by a factor of $H^{-4}$ at high aspect ratios. The cavity's trunk bandwidth is accordingly the most constrained of the three, using an anisotropic $\sigma=[\sigma_\xi,\sigma_\eta]$; Allen--Cahn and Burgers use a milder, isotropic bandwidth. Exact values for all three are given in Tables~\ref{tab:hyper-burgers}--\ref{tab:hyper}.

\subsection{Where the three instantiations differ}
\label{app:differences}
Beyond the three primary design choices outlined in Table~\ref{tab:axes}, the implementations differ across four specific algorithmic parameters (detailed in Appendices~\ref{app:distill} through \ref{app:hyperparams}): 

\textbf{(a)} The Burgers base operator employs a pure autograd-derived residual. Conversely, Allen--Cahn and the cavity utilize a weighted finite-difference stencil residual, which significantly reduces computational overhead for the biharmonic cavity operator and the batched Allen--Cahn Laplacian. 
\textbf{(b)} The Allen--Cahn framework incorporates a soft amplitude floor into the Part 1 loss, penalizing batch samples whose fields collapse toward $u{\equiv}0$. This serves as a complementary regularization to the seed lift $\ell_0$, mitigating the risk of the network encountering the competing trivial minimizer during early training iterations. 
\textbf{(c)} The extrapolation sequence detailed in \S\ref{sec:part2} requires one stage for the Burgers equation and the cavity's aspect-ratio variant (Appendix~\ref{app:extra}), two stages for the Allen--Cahn problem, and four stages for the principal cavity $\Real$-extrapolation. 
\textbf{(d)} Allen--Cahn and the primary $\Real$-extrapolated cavity deploy the ungated, piecewise-linear distillation schedule. Burgers and the cavity $H$-extrapolation use the physics-adaptive, reliability-gated schedule (\S\ref{sec:part3}).

\section{Distillation schedule and reliability gate}
\label{app:distill}

The Part~3 loss~\eqref{eq:distil-loss} weights the distillation term using two distinct components: a temporal scalar schedule $w(t)$ and an optional spatial per-point gate $\Gamma(\bx)$. The Allen--Cahn system and the cavity's $\Real$-extrapolation (main text) utilize an ungated, piecewise-linear temporal schedule ($\Gamma\equiv 1$, the identity); the Burgers equation and the cavity's $H$-extrapolation (Appendix~\ref{app:extra}) utilize a physics-adaptive temporal schedule coupled with the spatial reliability gate defined below.

\noindent\textbf{Piecewise-linear weight.} Over a fraction $f_{\min}$ of the total training budget $T$, $w(t)$ decays from an initial value $w_0$ to an intermediate value $w_{\min}$, and subsequently decays to $w_{\mathrm{end}}{=}0$ over the remainder of training:
\begin{equation}
    w(t) =
    \begin{cases}
        w_0 + (w_{\min}-w_0)\,\dfrac{t}{f_{\min}T}, & t \le f_{\min}T,\\[1.0em]
        w_{\min} + (w_{\mathrm{end}}-w_{\min})\,\dfrac{t-f_{\min}T}{(1-f_{\min})T}, & t > f_{\min}T.
    \end{cases}
\end{equation}
For the cavity's $\Real$-extrapolation ($\Real{=}3200$), the parameters are set to $(w_0, f_{\min}, w_{\mathrm{end}})=(0.7, 0.7, 0)$ with $w_{\min}{=}0$. This constitutes a linear decay to zero over the first $70\%$ of training, followed by exclusive residual minimization. Truncating $w(t)$ to exactly zero during the final training window allows the student network to refine the solution independent of the teacher's structural errors.

\noindent\textbf{Physics-adaptive weight.} While the piecewise-linear schedule is sufficient for Allen--Cahn and the cavity's $\Real$-extrapolation, we use a physics-adaptive schedule for the Burgers equation and the cavity's $H$-extrapolation to demonstrate the framework's flexibility in accommodating alternative distillation strategies. After an initial warm-up period of $T_{\mathrm{warm}}$ epochs, during which $w(t)$ increases linearly to $w_{\max}\alpha_w$, the weight dynamically tracks the convergence of the student network: $w = w_{\max}\alpha\cdot\min(1,\bar r_\phi/\bar r_\phi^{\,\mathrm{ref}})$. Here, $\bar r_\phi$ represents an exponential moving average (with a decay factor of $0.5$) of the student's PDE loss, and $\bar r_\phi^{\,\mathrm{ref}}$ is its reference value at the conclusion of the warm-up phase. The weight $w(t)$ is strictly zeroed once training surpasses a specified cutoff fraction of $T$. For the Burgers equation, the parameters are configured as $(\alpha_w, w_{\max}, T_{\mathrm{warm}}, \text{cutoff frac.}) = (0.8, 1.0, 200, 0.5)$; the cavity's $H$-extrapolation uses the same $(\alpha, w_{\max}, T_{\mathrm{warm}}, \text{cutoff frac.})$ values. This formulation dynamically attenuates the teacher's influence in response to the student's improving physics residual, rather than relying on a predetermined epoch schedule.

\paragraph{Spatial reliability gate $\Gamma(\bx)$.} 
When enabled (as in the Burgers and cavity $H$-extrapolation configurations), the spatial weight $\Gamma$ acts as a per-collocation-point factor that attenuates the distillation constraint in regions where the student's PDE residual $r_\phi(\bx)$ is large:
\begin{equation}
    \Gamma(\bx) \;=\; \big(1 - \tanh(\tau\, r_\phi(\bx))\big)\ \vee\ \Gamma_{\mathrm{floor}},
    \qquad \Gamma_{\mathrm{floor}} = 0.1.
\end{equation}
This value is clamped to a minimum floor $\Gamma_{\mathrm{floor}}$ to ensure the teacher's prior is never entirely deactivated. In regions where the student adequately satisfies the governing physics ($r_\phi$ is small), the gate approaches $1$ and the teacher's field is strictly enforced. Conversely, where the residual is large, the gate decays toward the floor value. For the Burgers equation, we utilize a lowered temperature of $\tau=2.0$ as the residual magnitude $|r_\phi|$ peaks inside the internal transition layer. Because this layer contains the critical structural prior provided by the 1D teacher, a high-temperature gate would inappropriately deactivate distillation in this region and restrict knowledge transfer exclusively to the trivial outer domains. Lowering $\tau$ ensures the gate remains sufficiently open at moderate residual values, a behavior we empirically verified by monitoring the gate-open fraction during training. The cavity's $H$-extrapolation instead uses $\tau=20.0$ (with the same floor $\Gamma_{\mathrm{floor}}=0.1$), reflecting the different residual scale of the two-dimensional momentum system.

\section{Hyperparameters and training details}
\label{app:hyperparams}

Table~\ref{tab:hyper-burgers} gives the Burgers hyperparameters (\S\ref{sec:results-burgers}), Table~\ref{tab:hyper-ac} the Allen--Cahn hyperparameters (\S\ref{sec:results-ac}), and Table~\ref{tab:hyper} the cavity hyperparameters for both the main-text $\Real$-extrapolation (\S\ref{sec:results-cavity}) and the Appendix~\ref{app:extra} $H$-extrapolation, which are independently trained base operators. All values are taken directly from the released configuration; code is released alongside the paper for further implementation details.

\begin{table}[t]
\centering
\small
\renewcommand{\arraystretch}{1.12}
\caption{Burgers hyperparameters (\S\ref{sec:results-burgers}).}
\label{tab:hyper-burgers}
\begin{tabular}{@{}p{0.5\linewidth}p{0.44\linewidth}@{}}
\toprule
\multicolumn{2}{@{}l}{\textit{Part 1 --- base operator}}\\
\midrule
Base box $(\nu, L)$                       & $[0.10,0.20]\times[1.0,2.0]$\\
Anchors                                   & $\{0.10,0.15,0.20\}\times\{1.0,1.5,2.0\}$ (9)\\
Trunk / branch / basis $K$                & $128\times5$ tanh / $128\times4$ tanh / 96\\
Fourier features $m$ / $\sigma$           & 64 / 6.0\\
Operator steps / LR ($\to$ min)           & $2\times10^4$ / $10^{-3}\!\to\!10^{-6}$\\
Collocation pts / param.\ batch / grad clip & 256 / 16 / 1.0\\
\midrule
\multicolumn{2}{@{}l}{\textit{Part 2 --- continuation}}\\
\midrule
Target                                    & $(\nu,L)=(0.05,4.0)$\\
Bounded-correction cap $\alpha$           & 0.63\\
$\beta$ range                             & $[10^{-3}, 10^{8}]$\\
Steps / LR / batch                        & $1.2\times10^4$ / $8\times10^{-4}$ / 8\\
\midrule
\multicolumn{2}{@{}l}{\textit{Part 3 --- guarded distillation}}\\
\midrule
Student / Fourier $m,\sigma$              & $128\times4$ tanh / $32, 8.0$\\
Epochs / LR ($\to$ min)                   & $4000$ / $5\times10^{-4}\!\to\!10^{-6}$\\
Collocation / BC / distil pts             & 1000 / 256 / 1000\\
BC weight / grad clip                     & 10.0 / 1.0\\
Distillation schedule                     & physics-adaptive + gate\\ Gate $\tau$/floor; $(\alpha_w,w_{\max},T_{\mathrm{warm}},\text{cutoff})$ & $2.0/0.1$; $(0.8,1.0,200,0.5)$\\
Seeds ($n$ / start)                       & 5 / 1729\\
\bottomrule
\end{tabular}
\end{table}

\begin{table}[t]
\centering
\small
\renewcommand{\arraystretch}{1.12}
\caption{Allen--Cahn hyperparameters (\S\ref{sec:results-ac}).}
\label{tab:hyper-ac}
\begin{tabular}{@{}p{0.5\linewidth}p{0.44\linewidth}@{}}
\toprule
\multicolumn{2}{@{}l}{\textit{Part 1 --- base operator}}\\
\midrule
Base box $(\lambda, H)$                   & $[35,80]\times[1.0,2.0]$\\
Anchors                                   & $\{35,50,65,80\}\times\{1.0,1.5,2.0\}$ (12), GN-polished\\
Trunk / branch / basis $K$                & $128\times4$ tanh / $128\times4$ tanh / 128\\
Fourier features $m$ / $\sigma$           & 32 / $[3.0,3.0]$\\
Operator steps / LR ($\to$ min)           & $3\times10^4$ / $10^{-3}\!\to\!10^{-6}$\\
FD residual grid / param.\ batch          & 48 / 8\\
Anti-collapse amplitude floor / weight    & $\mathrm{rms}(u)\ge0.25$ / $100$\\
GN iters / tol / damping / grad clip      & 60 / $10^{-10}$ / $10^{-4}$ / 1.0\\
\midrule
\multicolumn{2}{@{}l}{\textit{Part 2 --- continuation}}\\
\midrule
Targets                                   & two-rung ladder: $\lambda{=}18\to\lambda{=}12.5$ (both at $H{=}3$)\\
Bounded-correction cap $\alpha$           & 0.63\\
$\beta$ range                             & $[1.0, 2000]$\\
Poly.\ degree                             & 2\\
Steps / LR / batch                        & $1.2\times10^4$ / $5\times10^{-4}$ / 8\\
\midrule
\multicolumn{2}{@{}l}{\textit{Part 3 --- guarded distillation}}\\
\midrule
Student / Fourier $m,\sigma$              & $64\times2$ tanh / $32, 2.0$\\
Epochs / LR ($\to$ min)                   & $10^4$ / $5\times10^{-4}\!\to\!10^{-7}$\\
Collocation / BC / distil pts             & 2000 / 800 / 2000\\
BC weight / grad clip                     & 10.0 / 1.0\\
Distillation schedule                     & piecewise-linear\\ $(w_0,f_{\min},w_{\min},w_{\mathrm{end}})$ & $(1.0,\ 0.5,\ 0.1,\ 0.0)$\\
Seeds ($n$ / start)                       & 5 / 1729\\
\bottomrule
\end{tabular}
\end{table}

\begin{table}[t]
\centering
\small
\renewcommand{\arraystretch}{1.12}
\caption{Cavity hyperparameters: main-text $\Real$-extrapolation ($\Real{=}3200,H{=}1$, \S\ref{sec:results-cavity}) vs.\ the Appendix~\ref{app:extra} $H$-extrapolation ($H{=}3,\Real{=}400$).}
\label{tab:hyper}
\begin{tabular}{@{}p{0.4\linewidth}p{0.29\linewidth}p{0.25\linewidth}@{}}
\toprule
                                          & Main: $\Real{=}3200,H{=}1$ & App.\ E: $\Real{=}400,H{=}3$\\
\midrule
\multicolumn{3}{@{}l}{\textit{Part 1 --- base operator}}\\
Base box $(\Real, H)$                    & $[100,800]\times\{1.0\}$ & $[100,400]\times[1.0,2.0]$\\
Anchors                                  & 8 & 12\\
Trunk / branch / basis $K$               & $128\times8$ tanh / $128\times8$ tanh / 192 & $256\times6$ tanh / $256\times4$ tanh / 192\\
Fourier features $m$ / $\sigma$          & 64 / $[4.0,7.0]$ & 48 / $[4.0,7.0]$\\
Operator steps / LR ($\to$ min)          & $2\times10^5$ / $10^{-3}\!\to\!10^{-6}$ & $3\times10^5$ / $6\times10^{-4}\!\to\!10^{-6}$\\
FD residual grid / param.\ batch           & 160 / 4 & 64 / 8\\
Ridge $\mu$ / poly degree                & $10^{-2}$ / 2 & $10^{-2}$ / 3\\
Grad clip                                & 1.0 & 1.0\\
\midrule
\multicolumn{3}{@{}l}{\textit{Part 2 --- continuation}}\\
Targets                                  & 4-rung $\Real$ ladder: $1400{\to}2000{\to}2500{\to}3200$ & single-shot: $(\Real,H){=}(400,3)$\\
Bounded-correction cap $\alpha$          & 0.63 & 0.63\\
$\beta$ calibration clamp $[\beta_{\min},\beta_{\max}]$ & $[20, 2000]$ & $[20, 2000]$\\
$\beta$ within-step decay (start $\to$ end) & $\beta_0 \to \beta_0/20$ & $\beta_0 \to \beta_0/20$\\
Steps / LR / batch                       & $2.4\times10^4$ / $5\times10^{-4}$ / 8 & $2.4\times10^4$ / $5\times10^{-4}$ / 8\\
\midrule
\multicolumn{3}{@{}l}{\textit{Part 3 --- guarded distillation}}\\
Student / Fourier $m,\sigma$             & $256\times5$ tanh / $64, 6.0$ & $128\times3$ tanh / $32, 4.0$\\
Epochs / LR ($\to$ min)                  & $2.5\times10^5$ / $10^{-3}\!\to\!10^{-8}$ & $5\times10^4$ / $5\times10^{-4}\!\to\!10^{-6}$\\
Collocation / BC / distil pts            & 6000 / 4000 / 4000 & 4000 / 800 / 4000\\
Momentum / BC / gauge weight             & 1.0 / 10 / 10.0 & 1.0 / 10 / 1.0\\
Grad clip                                & 1.0 & 1.0\\
Distillation schedule                    & piecewise-linear & physics-adaptive + gate\\
Seeds ($n$ / start)                      & 5 / 1729 & single run\\
\bottomrule
\end{tabular}
\end{table}

\section{Reference-solver configurations}
\label{app:fd}

Reference fields are used only for post-hoc scoring and, for the cavity, for forming the eight main-text Part~1 anchors (twelve for the separate Appendix~\ref{app:extra} base operator); they never enter any training loss. For Burgers the reference is the Cole--Hopf closed form, evaluated exactly. For Allen--Cahn the reference is a Newton solve of the residual on a five-point Laplacian stencil. For the cavity the reference is a regularized-lid, streamfunction--vorticity finite-difference solver on a uniform $N_x{=}129$ grid (isoparametrically stretched to $N_y = \lceil (N_x - 1)H\rceil + 1$ for $H\neq1$), iterated to a steady state; it uses the same regularized lid profile $u_{\mathrm{lid}}$ as the operator (Appendix~\ref{app:cavity}). Each cavity anchor solve is gate-checked against the Marchi et al.~\citep{marchi2021lid} high-resolution benchmark where tabulated; the few-percent offset expected from the regularized-versus-singular lid is accounted for in that check.

\section{Additional results: aspect-ratio extrapolation}
\label{app:extra}

Section~\ref{sec:results-cavity} reports the cavity's Part 3 result at the $\Real$-extrapolated target ($\Real=3200$, $H=1$), reached from a base operator trained on $\Real\in[100,800]$ at fixed $H{=}1$ with $8$ anchors. Here we report a separate experiment stressing the complementary axis: a base operator trained on the full $(\Real,H)$ box with $12$ anchors (Table~\ref{tab:hyper}), single-shot extrapolated from $H\in[1,2]$ out to $H=3$ at fixed $\Real=400$, followed by Part 3 distillation of a fresh student PINN at that target. Every other mechanism is exactly as in the main pipeline (\S\ref{sec:part2}--\S\ref{sec:part3}). This example shows that the framework's benefit from distillation is not specific to the $\Real$-extrapolation axis reported in the main text; the same guarded-distillation mechanism accelerates convergence when the extrapolation instead stresses geometry.

\begin{figure}[h]
  \centering
  \includegraphics[width=0.8\linewidth]{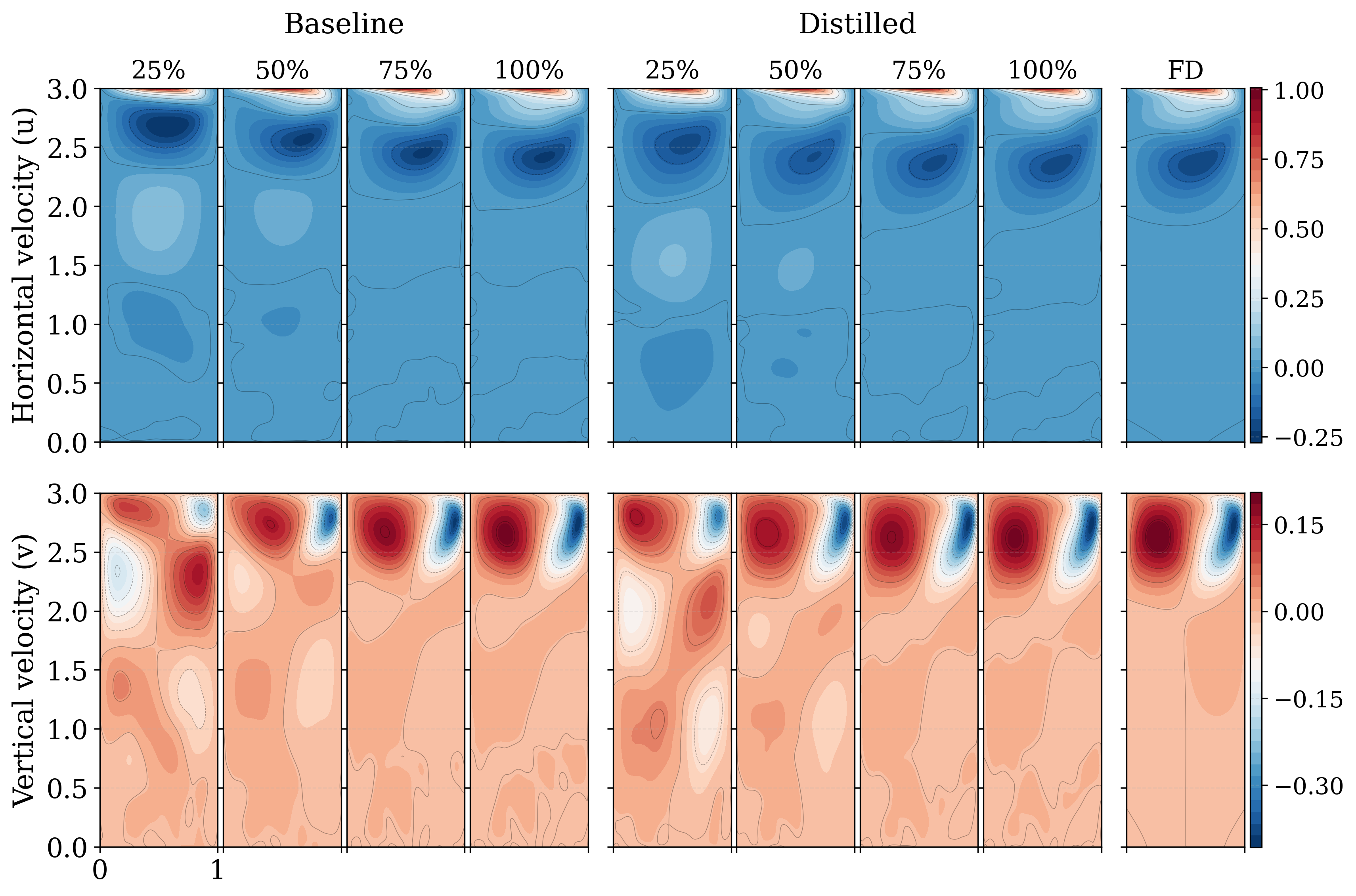}
  \caption{Lid-driven cavity, $H$-extrapolation example at $\mathrm{Re}=400$, $H=3$ (base box $H\in[1,2]$). Predicted horizontal ($u$, top row) and vertical ($v$, bottom row) velocity fields at 25/50/75/100\% of training for the baseline (left block) and distilled (right block) students, against the finite-difference reference (rightmost column).}
  \label{fig:ns2_contours}
\end{figure}

\begin{figure}[h]
  \centering
  \includegraphics[width=\linewidth]{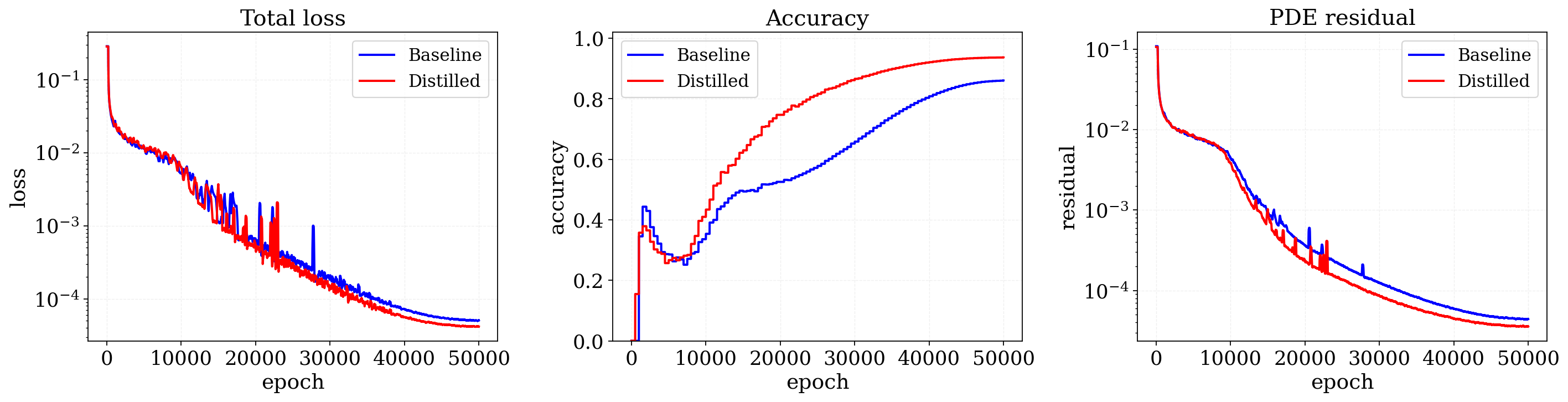}
  \caption{Lid-driven cavity, $H$-extrapolation example at $\mathrm{Re}=400$, $H=3$.}
  \label{fig:ns2_curves}
\end{figure}

Figure~\ref{fig:ns2_contours} compares the baseline and distilled students against the finite-difference reference across training progress (25/50/75/100\% of the training budget) for both velocity components. Figure~\ref{fig:ns2_curves} reports the corresponding total loss, accuracy against the FD reference, and PDE residual histories. As in the main-text $\Real=3200$ case, the distilled student converges faster and reaches a lower final residual and higher final accuracy than the baseline trained under an identical architecture and step budget.

\begin{figure}[htbp]
  \centering
  \includegraphics[width=0.8\linewidth]{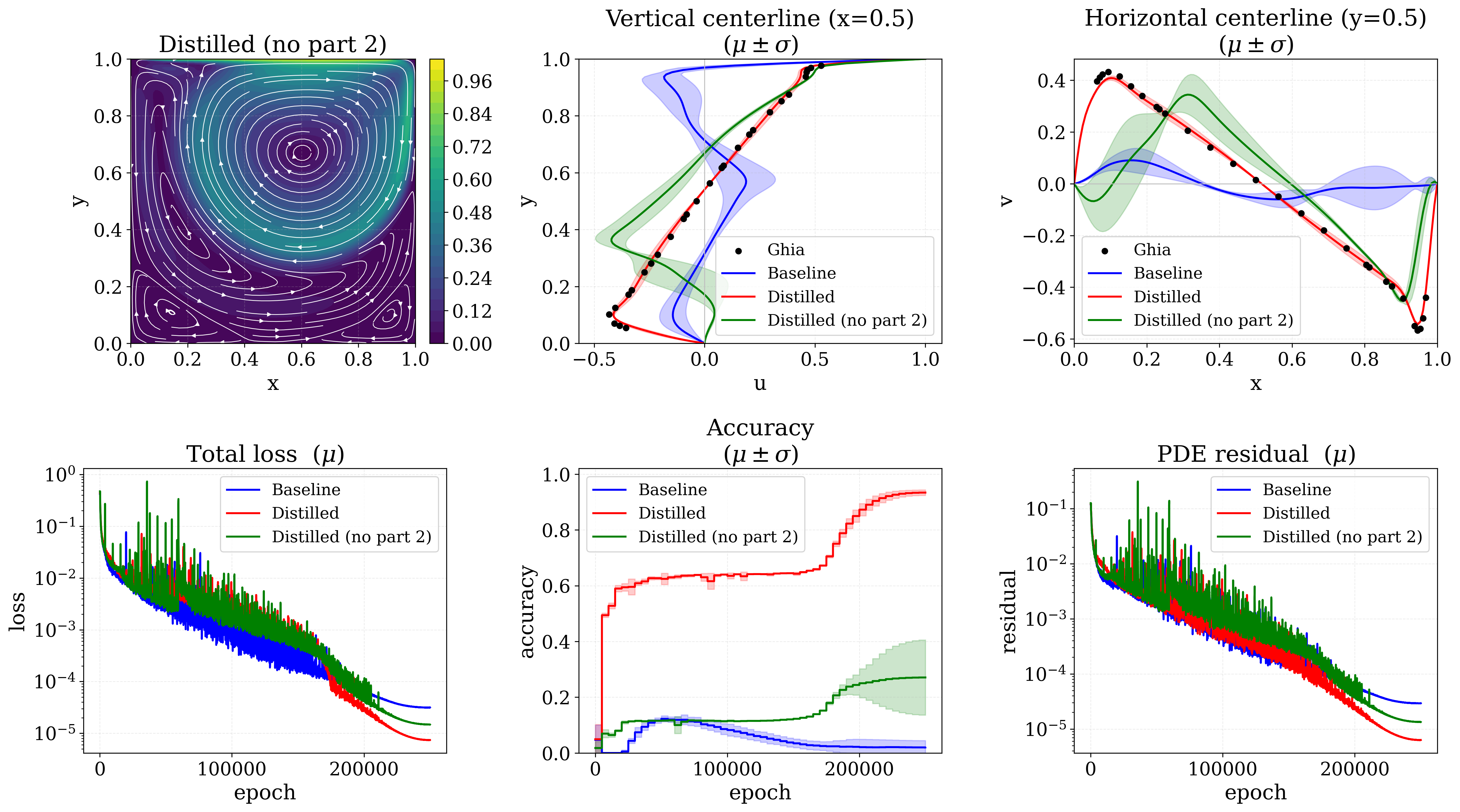}
  \caption{Cavity Part-2 removal ablation at $\mathrm{Re}=3200$, $H=1$. Solid lines and shaded regions represent $\mu \pm \sigma$ over $n=5$ seeds.}
  \label{fig:cavity-ablation}
\end{figure}


\begin{thebibliography}{41}
\providecommand{\natexlab}[1]{#1}
\providecommand{\url}[1]{\texttt{#1}}
\expandafter\ifx\csname urlstyle\endcsname\relax
  \providecommand{\doi}[1]{doi: #1}\else
  \providecommand{\doi}{doi: \begingroup \urlstyle{rm}\Url}\fi

\bibitem[Botella \& Peyret(1998)Botella and Peyret]{Botella1998Spectral}
O.~Botella and R.~Peyret.
\newblock Benchmark spectral results on the lid-driven cavity flow.
\newblock \emph{Computers \& Fluids}, 27\penalty0 (4):\penalty0 421--433, 1998.
\newblock \doi{10.1016/S0045-7930(98)00002-4}.

\bibitem[Cao et~al.(2024{\natexlab{a}})Cao, Goswami, and Karniadakis]{cao2024laplace}
Qianying Cao, Somdatta Goswami, and George~Em Karniadakis.
\newblock {Laplace neural operator for solving differential equations}.
\newblock \emph{Nature Machine Intelligence}, 6\penalty0 (6):\penalty0 631--640, 2024{\natexlab{a}}.

\bibitem[Cao et~al.(2024{\natexlab{b}})Cao, Goswami, Tripura, Chakraborty, and Karniadakis]{cao2024deep}
Qianying Cao, Somdatta Goswami, Tapas Tripura, Souvik Chakraborty, and George~Em Karniadakis.
\newblock {Deep neural operators can predict the real-time response of floating offshore structures under irregular waves}.
\newblock \emph{Computers \& Structures}, 291:\penalty0 107228, 2024{\natexlab{b}}.

\bibitem[Chakraborty(2021)]{chakraborty2021transfer}
Souvik Chakraborty.
\newblock Transfer learning based multi-fidelity physics informed deep neural network.
\newblock \emph{Journal of Computational Physics}, 426:\penalty0 109942, 2021.

\bibitem[Desai et~al.(2021)Desai, Mattheakis, Joy, Protopapas, and Roberts]{desai2021one}
Shaan Desai, Marios Mattheakis, Hayden Joy, Pavlos Protopapas, and Stephen Roberts.
\newblock One-shot transfer learning of physics-informed neural networks.
\newblock \emph{arXiv preprint arXiv:2110.11286}, 2021.

\bibitem[Erturk et~al.(2005)Erturk, Corke, and G{\"o}k{\c{c}}{\"o}l]{erturk2005numerical}
Ercan Erturk, Thomas~C Corke, and Cihan G{\"o}k{\c{c}}{\"o}l.
\newblock Numerical solutions of 2-d steady incompressible driven cavity flow at high reynolds numbers.
\newblock \emph{International journal for Numerical Methods in fluids}, 48\penalty0 (7):\penalty0 747--774, 2005.

\bibitem[Eshaghi et~al.(2026)Eshaghi, Anitescu, Valizadeh, Wang, Zhuang, and Rabczuk]{eshaghi2026nows}
Mohammad~Sadegh Eshaghi, Cosmin Anitescu, Navid Valizadeh, Yizheng Wang, Xiaoying Zhuang, and Timon Rabczuk.
\newblock Nows: Neural operator warm starts for accelerating iterative solvers.
\newblock \emph{Computer Methods in Applied Mechanics and Engineering}, 458:\penalty0 118989, 2026.

\bibitem[Ghia et~al.(1982)Ghia, Ghia, and Shin]{ghia1982high}
UKNG Ghia, Kirti~N Ghia, and CT~Shin.
\newblock High-re solutions for incompressible flow using the navier-stokes equations and a multigrid method.
\newblock \emph{Journal of computational physics}, 48\penalty0 (3):\penalty0 387--411, 1982.

\bibitem[Goswami et~al.(2020)Goswami, Anitescu, Chakraborty, and Rabczuk]{goswami2020transfer}
Somdatta Goswami, Cosmin Anitescu, Souvik Chakraborty, and Timon Rabczuk.
\newblock Transfer learning enhanced physics informed neural network for phase-field modeling of fracture.
\newblock \emph{Theoretical and Applied Fracture Mechanics}, 106:\penalty0 102447, 2020.

\bibitem[Goswami et~al.(2022)Goswami, Kontolati, Shields, and Karniadakis]{goswami2022deep}
Somdatta Goswami, Katiana Kontolati, Michael~D Shields, and George~Em Karniadakis.
\newblock Deep transfer operator learning for partial differential equations under conditional shift.
\newblock \emph{Nature Machine Intelligence}, 4\penalty0 (12):\penalty0 1155--1164, 2022.

\bibitem[Haghighat et~al.(2025)Haghighat, Adeli, Mousavi, and Juanes]{haghighat2025stonet}
Ehsan Haghighat, Mohammad~Hesan Adeli, S~Mohammad Mousavi, and Ruben Juanes.
\newblock Stonet: A neural operator for modeling solute transport in micro-cracked reservoirs.
\newblock \emph{Advances in Water Resources}, pp.\  105046, 2025.

\bibitem[Hinton et~al.(2015)Hinton, Vinyals, and Dean]{hinton2015distilling}
Geoffrey Hinton, Oriol Vinyals, and Jeff Dean.
\newblock Distilling the knowledge in a neural network.
\newblock \emph{arXiv preprint arXiv:1503.02531}, 2015.

\bibitem[Jagtap et~al.(2020)Jagtap, Kawaguchi, and Karniadakis]{jagtap2020adaptive}
Ameya~D Jagtap, Kenji Kawaguchi, and George~Em Karniadakis.
\newblock Adaptive activation functions accelerate convergence in deep and physics-informed neural networks.
\newblock \emph{Journal of Computational Physics}, 404:\penalty0 109136, 2020.

\bibitem[Karniadakis et~al.(2021)Karniadakis, Kevrekidis, Lu, Perdikaris, Wang, and Yang]{karniadakis2021physics}
George~Em Karniadakis, Ioannis~G Kevrekidis, Lu~Lu, Paris Perdikaris, Sifan Wang, and Liu Yang.
\newblock Physics-informed machine learning.
\newblock \emph{Nature Reviews Physics}, 3\penalty0 (6):\penalty0 422--440, 2021.

\bibitem[Kovachki et~al.(2023)Kovachki, Li, Liu, Azizzadenesheli, Bhattacharya, Stuart, and Anandkumar]{kovachki2023neural}
Nikola Kovachki, Zongyi Li, Burigede Liu, Kamyar Azizzadenesheli, Kaushik Bhattacharya, Andrew Stuart, and Anima Anandkumar.
\newblock Neural operator: Learning maps between function spaces with applications to pdes.
\newblock \emph{Journal of Machine Learning Research}, 24\penalty0 (89):\penalty0 1--97, 2023.

\bibitem[Krishnapriyan et~al.(2021)Krishnapriyan, Gholami, Zhe, Kirby, and Mahoney]{krishnapriyan2021characterizing}
Aditi Krishnapriyan, Amir Gholami, Shandian Zhe, Robert Kirby, and Michael Mahoney.
\newblock Characterizing possible failure modes in physics-informed neural networks.
\newblock \emph{Advances in neural information processing systems}, 34:\penalty0 26548--26560, 2021.

\bibitem[Li et~al.(2020{\natexlab{a}})Li, Kovachki, Azizzadenesheli, Liu, Bhattacharya, Stuart, and Anandkumar]{li2020fourier}
Zongyi Li, Nikola Kovachki, Kamyar Azizzadenesheli, Burigede Liu, Kaushik Bhattacharya, Andrew Stuart, and Anima Anandkumar.
\newblock Fourier neural operator for parametric partial differential equations.
\newblock \emph{arXiv preprint arXiv:2010.08895}, 2020{\natexlab{a}}.

\bibitem[Li et~al.(2020{\natexlab{b}})Li, Kovachki, Azizzadenesheli, Liu, Bhattacharya, Stuart, and Anandkumar]{li2020neural}
Zongyi Li, Nikola Kovachki, Kamyar Azizzadenesheli, Burigede Liu, Kaushik Bhattacharya, Andrew Stuart, and Anima Anandkumar.
\newblock Neural operator: Graph kernel network for partial differential equations.
\newblock \emph{arXiv preprint arXiv:2003.03485}, 2020{\natexlab{b}}.

\bibitem[Lu et~al.(2021{\natexlab{a}})Lu, Jin, Pang, Zhang, and Karniadakis]{lu2021learning}
Lu~Lu, Pengzhan Jin, Guofei Pang, Zhongqiang Zhang, and George~Em Karniadakis.
\newblock Learning nonlinear operators via deeponet based on the universal approximation theorem of operators.
\newblock \emph{Nature machine intelligence}, 3\penalty0 (3):\penalty0 218--229, 2021{\natexlab{a}}.

\bibitem[Lu et~al.(2021{\natexlab{b}})Lu, Pestourie, Yao, Wang, Verdugo, and Johnson]{lu2021physics}
Lu~Lu, Raphael Pestourie, Wenjie Yao, Zhicheng Wang, Francesc Verdugo, and Steven~G Johnson.
\newblock Physics-informed neural networks with hard constraints for inverse design.
\newblock \emph{SIAM Journal on Scientific Computing}, 43\penalty0 (6):\penalty0 B1105--B1132, 2021{\natexlab{b}}.

\bibitem[Lu et~al.(2022)Lu, Meng, Cai, Mao, Goswami, Zhang, and Karniadakis]{lu2022comprehensive}
Lu~Lu, Xuhui Meng, Shengze Cai, Zhiping Mao, Somdatta Goswami, Zhongqiang Zhang, and George~Em Karniadakis.
\newblock A comprehensive and fair comparison of two neural operators (with practical extensions) based on fair data.
\newblock \emph{Computer Methods in Applied Mechanics and Engineering}, 393:\penalty0 114778, 2022.

\bibitem[Mandl et~al.(2025)Mandl, Goswami, Lambers, and Ricken]{mandl2025separable}
Luis Mandl, Somdatta Goswami, Lena Lambers, and Tim Ricken.
\newblock Separable physics-informed deeponet: Breaking the curse of dimensionality in physics-informed machine learning.
\newblock \emph{Computer Methods in Applied Mechanics and Engineering}, 434:\penalty0 117586, 2025.

\bibitem[Marchi et~al.(2021)Marchi, Santiago, and Carvalho]{marchi2021lid}
Carlos~Henrique Marchi, Cosmo~Dami{\~a}o Santiago, and Carlos Alberto Rezende~de Carvalho, Jr.
\newblock Lid-driven square cavity flow: A benchmark solution with an 8192$\times$ 8192 grid.
\newblock \emph{Journal of Verification, Validation and Uncertainty Quantification}, 6\penalty0 (4):\penalty0 041004, 2021.

\bibitem[McClenny \& Braga-Neto(2023)McClenny and Braga-Neto]{mcclenny2023self}
Levi~D McClenny and Ulisses~M Braga-Neto.
\newblock Self-adaptive physics-informed neural networks.
\newblock \emph{Journal of Computational Physics}, 474:\penalty0 111722, 2023.

\bibitem[Perlin(2002)]{perlin2002improving}
Ken Perlin.
\newblock Improving noise.
\newblock In \emph{Proceedings of the 29th annual conference on Computer graphics and interactive techniques}, pp.\  681--682, 2002.

\bibitem[Raissi et~al.(2019)Raissi, Perdikaris, and Karniadakis]{raissi2019physics}
Maziar Raissi, Paris Perdikaris, and George~E Karniadakis.
\newblock Physics-informed neural networks: A deep learning framework for solving forward and inverse problems involving nonlinear partial differential equations.
\newblock \emph{Journal of Computational physics}, 378:\penalty0 686--707, 2019.

\bibitem[Sirignano \& Spiliopoulos(2018)Sirignano and Spiliopoulos]{sirignano2018dgm}
Justin Sirignano and Konstantinos Spiliopoulos.
\newblock Dgm: A deep learning algorithm for solving partial differential equations.
\newblock \emph{Journal of computational physics}, 375:\penalty0 1339--1364, 2018.

\bibitem[Tancik et~al.(2020)Tancik, Srinivasan, Mildenhall, Fridovich-Keil, Raghavan, Singhal, Ramamoorthi, Barron, and Ng]{tancik2020fourier}
Matthew Tancik, Pratul Srinivasan, Ben Mildenhall, Sara Fridovich-Keil, Nithin Raghavan, Utkarsh Singhal, Ravi Ramamoorthi, Jonathan Barron, and Ren Ng.
\newblock Fourier features let networks learn high frequency functions in low dimensional domains.
\newblock \emph{Advances in neural information processing systems}, 33:\penalty0 7537--7547, 2020.

\bibitem[Tripura \& Chakraborty(2023)Tripura and Chakraborty]{tripura2023wavelet}
Tapas Tripura and Souvik Chakraborty.
\newblock Wavelet neural operator for solving parametric partial differential equations in computational mechanics problems.
\newblock \emph{Computer Methods in Applied Mechanics and Engineering}, 404:\penalty0 115783, 2023.

\bibitem[Wang et~al.(2021{\natexlab{a}})Wang, Teng, and Perdikaris]{wang2021understanding}
Sifan Wang, Yujun Teng, and Paris Perdikaris.
\newblock Understanding and mitigating gradient flow pathologies in physics-informed neural networks.
\newblock \emph{SIAM Journal on Scientific Computing}, 43\penalty0 (5):\penalty0 A3055--A3081, 2021{\natexlab{a}}.

\bibitem[Wang et~al.(2021{\natexlab{b}})Wang, Wang, and Perdikaris]{wang2021eigenvector}
Sifan Wang, Hanwen Wang, and Paris Perdikaris.
\newblock On the eigenvector bias of fourier feature networks: From regression to solving multi-scale pdes with physics-informed neural networks.
\newblock \emph{Computer Methods in Applied Mechanics and Engineering}, 384:\penalty0 113938, 2021{\natexlab{b}}.

\bibitem[Wang et~al.(2021{\natexlab{c}})Wang, Wang, and Perdikaris]{wang2021learning}
Sifan Wang, Hanwen Wang, and Paris Perdikaris.
\newblock Learning the solution operator of parametric partial differential equations with physics-informed deeponets.
\newblock \emph{Science advances}, 7\penalty0 (40):\penalty0 eabi8605, 2021{\natexlab{c}}.

\bibitem[Wang et~al.(2022)Wang, Yu, and Perdikaris]{wang2022and}
Sifan Wang, Xinling Yu, and Paris Perdikaris.
\newblock When and why pinns fail to train: A neural tangent kernel perspective.
\newblock \emph{Journal of Computational Physics}, 449:\penalty0 110768, 2022.

\bibitem[Wang et~al.(2023{\natexlab{a}})Wang, Sankaran, Wang, and Perdikaris]{wang2023expert}
Sifan Wang, Shyam Sankaran, Hanwen Wang, and Paris Perdikaris.
\newblock An expert's guide to training physics-informed neural networks.
\newblock \emph{arXiv preprint arXiv:2308.08468}, 2023{\natexlab{a}}.

\bibitem[Wang et~al.(2024{\natexlab{a}})Wang, Li, Chen, and Perdikaris]{Wang2024PirateNets}
Sifan Wang, Bowen Li, Yuhan Chen, and Paris Perdikaris.
\newblock Piratenets: Physics-informed deep learning with residual adaptive networks.
\newblock \emph{Journal of Machine Learning Research}, 25\penalty0 (402):\penalty0 1--51, 2024{\natexlab{a}}.

\bibitem[Wang et~al.(2024{\natexlab{b}})Wang, Sankaran, and Perdikaris]{wang2024respecting}
Sifan Wang, Shyam Sankaran, and Paris Perdikaris.
\newblock Respecting causality for training physics-informed neural networks.
\newblock \emph{Computer Methods in Applied Mechanics and Engineering}, 421:\penalty0 116813, 2024{\natexlab{b}}.

\bibitem[Wang et~al.(2026)Wang, Wong, Ruan, and Goswami]{wang2026causality}
Wei Wang, Tang~Paai Wong, Haihui Ruan, and Somdatta Goswami.
\newblock Causality-respecting adaptive refinement for pinns: enabling precise interface evolution in phase field modeling.
\newblock \emph{Machine Learning for Computational Science and Engineering}, 2\penalty0 (1):\penalty0 10, 2026.

\bibitem[Wang et~al.(2023{\natexlab{b}})Wang, Meng, Jiang, Xiang, and Karniadakis]{wang2023solution}
Zhicheng Wang, Xuhui Meng, Xiaomo Jiang, Hui Xiang, and George~Em Karniadakis.
\newblock Solution multiplicity and effects of data and eddy viscosity on navier-stokes solutions inferred by physics-informed neural networks.
\newblock \emph{arXiv preprint arXiv:2309.06010}, 2023{\natexlab{b}}.

\bibitem[Weiss et~al.(2016)Weiss, Khoshgoftaar, and Wang]{weiss2016survey}
Karl Weiss, Taghi~M Khoshgoftaar, and DingDing Wang.
\newblock A survey of transfer learning.
\newblock \emph{Journal of Big data}, 3\penalty0 (1):\penalty0 9, 2016.

\bibitem[Wu et~al.(2023)Wu, Zhu, Tan, Kartha, and Lu]{wu2023comprehensive}
Chenxi Wu, Min Zhu, Qinyang Tan, Yadhu Kartha, and Lu~Lu.
\newblock A comprehensive study of non-adaptive and residual-based adaptive sampling for physics-informed neural networks.
\newblock \emph{Computer Methods in Applied Mechanics and Engineering}, 403:\penalty0 115671, 2023.

\bibitem[Yu et~al.(2022)Yu, Lu, Meng, and Karniadakis]{yu2022gradient}
Jeremy Yu, Lu~Lu, Xuhui Meng, and George~Em Karniadakis.
\newblock Gradient-enhanced physics-informed neural networks for forward and inverse pde problems.
\newblock \emph{Computer Methods in Applied Mechanics and Engineering}, 393:\penalty0 114823, 2022.

\end{thebibliography}
\end{document}